\documentclass[10pt,conference,a4paper]{IEEEtran}

\usepackage{cite}

\usepackage[pdftex]{graphicx}
\graphicspath{{../pdf/}{../jpeg/}}
\DeclareGraphicsExtensions{.pdf,.jpeg,.png}

\usepackage[lofdepth,lotdepth]{subfig}

\usepackage{textcomp}
\usepackage{enumerate}

\usepackage{amsmath,amssymb,amsfonts,latexsym,amscd,epsf,psfrag} 
\usepackage{stackrel}

\def\bbbr{{\mathbb R}} 
\def\bbbz{{\mathbb Z}}
\def\calD{\mathcal{D}}
\def\calT{\mathcal{T}}
\def\calS{\mathcal{S}}

\usepackage{amsmath}
\usepackage{amsfonts}
\usepackage{amssymb}

\newcommand{\local}{\operatorname{local}}

\newcommand{\avg}{\operatorname{avg}}
\newcommand{\conc}{\operatorname{conc}}

\newcommand{\diag}{\operatorname{diag}}
\graphicspath{{images-janlin19/}{}}

\begin{document}
\title{Exploring the ability of CNNs to generalise to previously unseen scales over wide scale ranges	\thanks{Shortened version in International Conference on Pattern Recognition
		(ICPR 2020), pages 1181-1188, Jan 2021. The support from the Swedish Research Council 
		(contract 2018-03586) is gratefully acknowledged.}
%	\thanks{This paper is accepted for publication at ICPR2020.}	
}

\author{\IEEEauthorblockN{Ylva Jansson and Tony Lindeberg} \IEEEauthorblockA{Computational Brain Science Lab, Division of Computational Science and Technology\\ KTH Royal Institute of Technology, Stockholm, Sweden}}

\IEEEoverridecommandlockouts



\maketitle


\begin{abstract}
	The ability to handle large scale variations is crucial for many real world visual tasks. 
	A straightforward approach for handling scale in a deep network is to process an image at several scales simultaneously in a set of \emph{scale channels}. Scale invariance can then, in principle, be achieved by using weight sharing between the scale channels together with max or average pooling over the outputs from the scale channels.
	The ability of such 
	\emph{scale channel networks} to generalise to scales not present in the training set over significant scale ranges has, however, not previously been explored. 
	We, therefore, present a theoretical analysis of invariance and covariance properties of scale channel networks and perform an experimental evaluation of the ability of different types of scale channel networks to generalise to previously unseen scales. 
	We identify limitations of previous approaches and propose a new type of \emph{foveated scale channel architecture}, where the scale channels process increasingly larger parts of the image with decreasing resolution. 
Our proposed FovMax and FovAvg networks perform almost identically over a scale range of 8, also when training on \emph{single scale training data}, and do also give improvements in the small sample regime.	
	%A straightforward approach to handling scale in a deep neural network is to process multiple rescaled image copies in a set of scale channels (subnetworks).	
\end{abstract}


% \fi
\IEEEpeerreviewmaketitle

\section{Introduction}
\label{sec:intro}

Scaling transformations are as pervasive in natural image data as translations. In any natural scene, the size of the projection of an object on the retina or a digital sensor varies continuously with the distance between the object and the observer. 
Convolutional neural networks (CNNs) already encode structural assumptions about translation invariance and locality.  
A vanilla CNN is, however, not designed for multi-scale processing, since the fixed size of the filters together with the depth and max-pooling strategy applied implies a preferred scale.
Encoding structural priors about visual transformations, including scale or affine invariance, is an integrated part of a range of successful classical computer vision approaches. There is also a growing body of work on invariant CNNs, especially concerning invariance to 2D/3D rotations and flips (see e.g. \cite{BruMal-TPAMI2013,CohWel-ICML2016,LapSavBuh-CVPR2016}). 
The possibilities for CNNs to generalise to previously unseen scales have, however, not been well explored. We propose that structural assumptions about scale could, similarly to translation covariance, be a useful prior in convolutional neural networks. 
\emph{Scale-invariant CNNs} could enable both multi-scale processing and predictable behaviour when encountering objects at novel scales, without the need to fully span all possible scales in the training set. 

One of the simplest CNN architectures used for covariant and invariant image processing is a channel network (also referred to as siamese network) \cite{LapSavBuh-CVPR2016}.
%DieWilDam-NRAS2015 
In such an architecture, transformed copies of the input image are processed in parallel by different ``channels" (subnetworks) corresponding to a set of image transformations. If combined with weight sharing and max or average pooling over the output from the channels, this approach can enable invariant recognition for finite transformation groups. 

An \emph{invariant scale channel network} is a natural extension of invariant channel networks for rotations  \cite{LapSavBuh-CVPR2016}.  It can equivalently be seen as a way of extending ideas underlying the classical scale-space methodology \cite{Iij62-TR,Wit83,Koe84-BC,lindeberg1993-scspbook,lindeberg1994-scsparticle,Flo97-book,scalespaceJapan_weickert99,haarromeny-04book} to deep learning. It should be noted that a channel architecture for scale-invariant recognition poses additional challenges compared to recognition over finite groups. First, scaling transformations are, as opposed to 2D or 3D rotations, not a compact group (intuitively, there is no smallest or largest scale). Second, scaling implies a change in image size and resolution for discrete image data. 
The subject of this paper is to investigate the possibility for CNNs to generalise to previously unseen scales by means of a scale channel architecture. 

\subsection{Contribution and novelty}
The key contributions of our work are as follows:
\begin{itemize}
\item We perform a theoretical analysis of invariance and covariance properties of scale channel networks.
\item We present a new family of invariant foveated scale channel networks. %, where the individual scale channels process increasingly larger parts of the image with decreasing resolution. %combine
%. Together with max and average pooling, this leads to our proposed FovMax and FovAvg networks. 
%\item We construct a 
\item We evaluate different types of scale channel networks and a standard CNN on the task of \emph{scale generalisation over wide scale ranges}, using a new variation of the MNIST dataset with large scale variations.
\item We demonstrate inherent limitations of previous scale channel approaches.
\item We show that our proposed foveated networks can enable very good generalisation to unseen scales and improvements in the small sample regime. 
%  two previously used scale channel network designs/\allowbreak methodologies do, in fact, not generalise well to scales not present in the training set, while the proposed 
% when the image resolution is sufficient and there is an absence of boundary effects. 
%\item 
%We, thus, present a proof of concept that generalisation over wide scale ranges is, indeed, possible in CNNs.
\end{itemize}
This is, to our knowledge, the first study to evaluate and demonstrate means for CNNs to \emph{generalise to unseen scales over significant scale ranges}.

\subsection{Related work}
In classical scale-space theory \cite{Iij62-TR,Wit83,Koe84-BC,lindeberg1993-scspbook,lindeberg1994-scsparticle,Flo97-book,scalespaceJapan_weickert99,haarromeny-04book},
a multi-scale representation of an input image is created by convolving the image with a set of rescaled Gaussian kernels and Gaussian derivative filters, which are then often combined in non-linear ways. The scale channel networks described in this paper  can be seen as an extension of this philosophy of 
 processing an image \emph{at all scales simultaneously}, but using deep non-linear feature extractors learned from data.

CNNs can give impressive performance but they are sensitive to scale variations. Performance degrades for scales not present in the training set \cite{EngTsiSchMad-ICML2019,FawFro-BMVC2015,SinDav-CVPR2018}, different network structure is optimal for small vs large images \cite{SinDav-CVPR2018} and it is possible to construct adversarial examples by means of small translations rotations and scalings \cite{EngTsiSchMad-ICML2019,FawFro-BMVC2015}.
State-of-the-art CNN based object detection approaches 
all employ different mechanisms to deal with scale variability, e.g. branching off classifiers at different depths \cite{CaiFanFerVas-ECCV2016}, learning to transform the input or the filters \cite{JadSimZisKav-NIPS2015,YuKol-ICLR2016}, or using different types of image pyramids \cite{SerEigZha-arXiv2013,Gir-ICCV2015,LinDolGirhe-CVPR2017}.
The goal of these approaches has, however, not been to generalise to \emph{previously unseen scales} and they lack the structure necessary for true scale invariance. 

Examples of handcrafted scale invariant hierarchical descriptors are \cite{SifMal-CVPR2013,Lin2020provably}. We are, here, interested in combining scale invariance with learning. There exist some previous work aimed explicitly at scale invariant recognition in CNNs  \cite{XuXiaZhaYan-arXiv2014,KanShaJac-arXiv2014,MarKelLob-arXiv2018,WorWel-NIPS2019}. These approaches have, however, either not been evaluated for the task of generalisation to scales \emph{not present in the training set} \cite{KanShaJac-arXiv2014,MarKelLob-arXiv2018,WorWel-NIPS2019} or only across a very limited scale range \cite{XuXiaZhaYan-arXiv2014}. Previous scale channel networks exist, but are explicitly designed for multi-scale processing \cite{FarCouNajLeC-2013,NooPos-PR2016} rather than scale invariance or have not been evaluated with regard to their ability to generalise to unseen scales over any significant scale range \cite{SerEigZha-arXiv2013,XuXiaZhaYan-arXiv2014}. 

\section{Theory}
\label{sec:theory}
%%%%%%%%%%%%%%%%%%%%%%%%%%%%%%%%%%%%%%%%%
% Invariance and covariance
%%%%%%%%%%%%%%%%%%%%%%%%%%%%%%%%%%%%%%%%%
In this section, we will introduce a mathematical framework for scale channel networks based on a continuous model of the image space. This model enables straightforward analysis of the covariance and invariance properties of the channel networks that are later approximated in a discrete implementation. We, here, generalise previous analysis of invariance properties of channel networks \cite{LapSavBuh-CVPR2016} to scale channel networks. We further analyse covariance properties and additional options for aggregating information across transformation channels.

\subsection{Images and image transformations}
We consider images $f: \mathbb{R}^N \to \mathbb{R}$ that are measurable functions in $L_\infty(\bbbr^N)$ and denote this space of images as $V$. 
A \emph{group of image transformations} corresponding to a group $G$ is a family of image transformations $\calT_g$ ($g \in G$) with a group structure. %, i.e. fulfilling the group axioms of closure, identity, associativity and inverse.
% (for a review of basic group theory see [ref]). 
We denote the combination of two group elements $g,h \in G$ by $gh$ and the cardinality of $G$ as $|G|$. 
Formally, a group $G$ induces an \emph{action on functions} by acting on the underlying space on which the function is defined (here the image domain). We are here  interested in the group of \emph{uniform scalings} around $x_0$ with the group action
\begin{align}
(\calS_{s, x_0} f ) (x') &= f(x) \text{,~~~~} x' = S_s(x - x_0) + x_0 
\label{eq:scale-def},
\end{align}
where $S_s = \diag(s)$. For simplicity, we often assume $x_0=0$  and denote $\calS_{s,0}$ as $\calS_s$ corresponding to 
\begin{equation}
(\calS_s f)(x) = f(S_s^{-1} x) = f_s(x).
\label{eq:scale-def2}
\end{equation}
We will also consider the translation group with the action (where $\delta \in \bbbr^N$)
\begin{align}
(\calD_\delta f) (x') &= f(x) \text{,~~~~}  x' = x + \delta. % \text{~~~~~~~}  x' = x - t
\label{eq:trans-def}
\end{align}

\subsection{Invariance and covariance}
Consider a general feature extractor $\Lambda: V \to \mathbb{K}$ that maps an image $f \in V$ to a feature representation $y \in \mathbb{K}$. 
In our continuous model, $\mathbb{K}$ will typically correspond to a set of $M$  feature maps (functions) so that $\Lambda f \in V^M$.
%$c \in \{1,2 \cdots M\}$
This is a continuous analogue of a discrete convolutional feature map with $M$ features.

A feature extractor $\Lambda$ is \emph{covariant} to a transformation group $G$ (formally to the group action) if there exists an \emph{input independent} transformation $\tilde{\calT}_g$ that can align the feature maps of a transformed image with those of the original image
%. That is i.e. there exists an operator such that
\begin{equation}
\Lambda (\calT_g f) = \tilde{\calT}_g (\Lambda f)  
\label{eq:covariance}
\end{equation}
for all $g \in G$ and $f \in V$.
Thus, for a covariant feature extractor it is possible to predict the feature maps of a transformed image from the feature maps of the original image. 

A feature extractor $\Lambda$ is \emph{invariant} to a transformation group $G$ if the feature representation of a transformed image is \emph{equal to} the feature representation of the original image 
\begin{equation}
\Lambda (\calT_g f) = \Lambda (f) % \text{~~~~~~~}
\label{eq:invariance}
\end{equation}
for all $g \in G$ and $f \in V$.
Invariance is thus a special case of covariance where $\tilde{\calT_g}$ is the identity transformation. 

\subsection{Continuous model of a CNN}

Let $\phi:  V \to V^{M_k}$ denote a continuous CNN with $k$ layers and $M_i$ feature channels in layer $i$.
Let $\theta^{(i)}$ represent the transformation between layers $i-1$ and $i$ such that
\begin{align}
(\phi^{(i)} f)(x,c) &= (\theta^{(i)} \theta^{(i-1)} \cdots \theta^{(2)} \theta^{(1)} f)(x,c), 
\label{eq:phi_i-def}
\end{align} 
where $c \in \{1,2, \dots M_k\}$ denotes the feature channel and $\phi = \phi^{(k)}$.
We model the transformation $\theta^{(i)}$ between two adjacent layers $\phi^{(i-1)}f$ and $\phi^{(i)}f$ as a convolution followed by the addition of a bias term $b_{i,c} \in \bbbr$ and the application of a pointwise non-linearity $\sigma_i:\bbbr \to \bbbr$:

\begin{multline}
(\phi^{(i)} f)(x, c)\\ =  \sigma_i \left( \sum_{m=1}^{M_{i-1}} \int_{\xi \in \bbbr^N } (\phi^{(i-1)}f)(x-\xi, m)\, g^{(i)}_{m,c}(\xi) \, d\xi + b_{i,c}
\right)
\label{eq:phi_integral}
\end{multline}
where $g^{(i)}_{m,c} \in L_1(\bbbr^N)$
denotes the convolution kernel that propagates information from feature channel $m$ in layer $i-1$ to output feature channel $c$ in layer $i$.
A final fully connected classification layer with compact support can also be modelled as a convolution combined with a non-linearity $\sigma_k$ that represents \emph{a softmax operation} over the feature channels.

\begin{figure*}[hbpt]
	\begin{center}
		\includegraphics[width=0.8\textwidth]{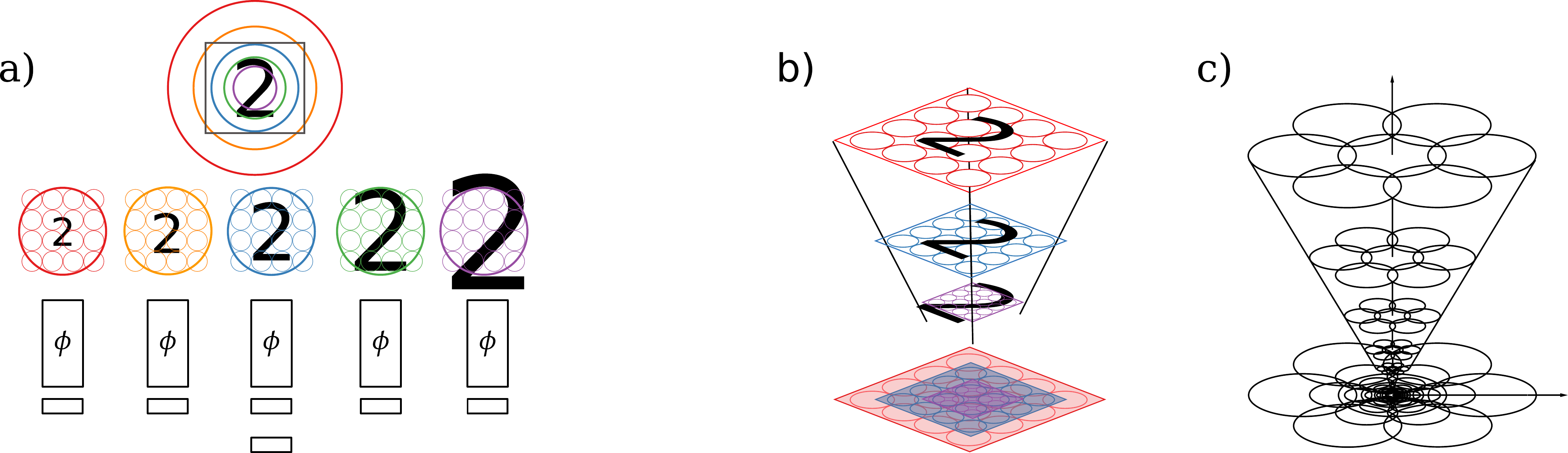} 
	\end{center}
	\caption{\emph{Foveated scale channel networks.} a) 
		Foveated scale channel network that process an image of the digit 2. Since each scale channel has a fixed size receptive field/support region in the scale channels, they will together process input regions corresponding to varying sizes in the original image (circles of corresponding colors).
		b) This corresponds to a type of foveated processing, where the center of the image is processed with high resolution, which works well to detect small objects, while larger regions are processed using gradually reduced resolution, which enables detection of larger objects. c) There is a close similarity between this model and the foveal scale space model \cite{CVAP166}, which was motivated by a combination of regular scale space axioms with a complementary assumption of a uniform limited processing capacity at all scales.
	}
	\label{fig:foveated-processing}
\end{figure*}

\subsection{Scale channel networks}
\label{sec:scale-channel-networks}
The key idea underlying \emph{channel networks} is to process transformed copies of an input image in parallel, in a set of network ``channels" (subnetworks) with shared weights. For finite transformation groups, such as discrete rotations, using one channel corresponding to each group element and applying max pooling over the channel dimension can give an invariant output code. For continuous but compact groups, invariance can instead be achieved for a discrete subgroup. 

The scaling group does, however, imply additional challenges, since it is neither finite nor compact. The key question that we address here, is whether a scale channel network can still support invariant recognition. 

We will define a multi-column \emph{scale channel network} 
$\Lambda: V \to V^{M_k}$
for the group of scaling transformations $S$ by using a single base network $\phi: V \to V^{M_k}$ to define a set of \emph{scale channels} $\{\phi_s \}_{s \in S}$
\begin{equation}
(\phi_s f)(x,c) = (\phi\, \calS_{s} f)(x,c) = (\phi f_s)(x,c),
\label{eq:phi_s-def}
\end{equation}
where each channel thus applies exactly the same operation to a scaled copy of the input image (see Figure \ref{fig:foveated-processing}a). 
We will denote the mapping from the input image to the scale channel feature maps at depth $i$ as $\Gamma^{(i)}: V \to V^{M_i |S|}$
\begin{equation}
(\Gamma^{(i)} f)(x,c,s) =  (\phi^{(i)}_s f)(x, c)  = (\phi^{(i)} \calS_s f)(x,c).
\label{eq:gamma_s-def}
\end{equation}
A scale channel network invariant to the continuous group of uniform scaling transformations $S = \{s \in \bbbr_+ \}$ 
can be constructed using an \emph{infinite} set of scale channels $\{ \phi_s \}_{s \in S}$. The following analysis also holds for a set of scale channels corresponding to a discrete subgroup of the group of uniform scaling transformations such that 
$S = \{\gamma^i |  i \in \bbbz\}$, $\gamma > 1$.

The final output $\Lambda f$ from the  scale channel network is an aggregation across the scale dimension of the last layer scale channel feature maps. In our theoretical treatment, we combine the output of the scale channels by the supremum
\begin{equation}
(\Lambda_{\sup} f)(x,c) = \sup_{s \in S} \left[  ( \phi_s f)(x,c,s) \right]
\label{eq:lambda_s-def}.
\end{equation}
Other permutation invariant operators such as averaging operations, could also be used. For this construction, the network output will be invariant to \emph{rescalings around $x_0=0$ for all $x$} such that $(\Lambda_{\sup} f)(x,c) = (\Lambda_{\sup} S_{s}f)(x, c)$ (global scale invariance). This architecture is appropriate  when characterising a single centered object that might vary in scale and it is the main architecture we explore in this paper. 
Alternatively, one may instead pool over \emph{corresponding image points} in the original image by operations of the form
\begin{equation}
\label{eq:lambda_s-skewed-def}
(\Lambda_{\sup}^{\local} f)(x, c) = \sup_{s \in S} \{ (\phi_s f)(S_s x, c) \}
\end{equation}
This descriptor instead has the invariance property
$(\Lambda_{\sup}^{\local} f)(x_0, c) = (\Lambda_{\sup}^{\local} S_{s, x_0}f)(x_0, c)$ \emph{for all $x_0$}, i.e. when scaling around an arbitrary image point, the output \emph{at that specific point} does not change (local scale invariance). This property makes it more suitable to describe scenes with multiple objects. 

\subsubsection{Scale covariance}
Consider a scale channel network $\Lambda$ (\ref{eq:lambda_s-def}) that expands the input over the group of uniform scaling transformations $S$. We can relate the feature map representation $\Gamma^{(i)}$ %(\ref{eq:gamma_s-def}) 
for a scaled image copy $\calS_t f$ for $t \in S$ and its original $f$ in terms of operator notation as
\begin{align}
&(\Gamma^{(i)} \calS_t f)(x,c,s) = (\phi_s^{(i)} \, \calS_t f)(x,c) \nonumber\\ 
&=(\phi^{(i)} \, \calS_s \, \calS_t f)(x,c) = (\phi^{(i)} \, \calS_{s t} f)(x,c) \nonumber\\ 
&= (\phi_{st}^{(i)} f)(x,c) = (\Gamma^{(i)} f)(x,c, s t),
\label{eq:scale-covariance}
\end{align}
where we have used the definitions (\ref{eq:phi_s-def}) and (\ref{eq:gamma_s-def}) together with the fact that $S$ is a group. A scaling of an image thus only results in a multiplicative shift in the scale dimension of the feature maps. A more general and more rigorous proof using an integral representation of a scale channel network is given in Section \ref{sec:invariance-covariance-proof}. 

\subsubsection{Scale invariance}
\label{sec:scale-invariance}
Consider the scale channel network $\Lambda_{\sup}$ (\ref{eq:lambda_s-def}) that selects the supremum over scales. We will show that $\Lambda_{\sup}$ is scale invariant i.e. that
\begin{equation}
(\Lambda_{\sup}\, \calS_t f)(x,c) = (\Lambda_{\sup} f)(x,c).
\label{eq:scale-invariance}
\end{equation}
First, (\ref{eq:scale-covariance}) gives $\{\phi^{(i)}_s (\calS_t f)\}_{s \in S} = \{ \phi_{st}^{(i)} (f) \}_{s \in S}$. Then, we note that $\{st\}_{s \in S} = St = S$. 
This holds both in the case when $S = \bbbr_+$ and in the case when $S = \{\gamma^i |  i \in \bbbz\}$.
Thus, we have
\begin{multline} 
\{(\phi_s^{(i)} \calS_t f)(x,c)\}_{s \in S}
= \{(\phi^{(i)}_{s t} f)(x,c)\}_{s \in S}\\ =  \{(\phi^{(i)}_{s} f)(x,c)\}_{s \in S}
\label{eq:phi_maps_scale},
\end{multline}
i.e. \emph{the set} of outputs from the scale channels for a transformed image is equal to the set of outputs from the scale channels for its original image. 
For any permutation invariant aggregation operator, such as the supremum, we have that
\begin{multline}
(\Lambda_{\sup}\, \calS_s f)(x,c) = \sup_{s \in S} \{(\phi^{(k)}_{s t} f)(x, c)\} \\
=   \sup_{s \in S} \{(\phi^{(k)}_{s} f)(x,c)\}  = (\Lambda_{\sup} f)(x, c),
\label{eq:lambda-scale-invariant}
\end{multline} 
and, thus, $\Lambda$ is invariant to uniform rescalings. % around $x = 0$. 

\subsection{Proof of scale and translation covariance using an integral representation of a scale channel network}
\label{sec:invariance-covariance-proof}
We, here, prove the transformation property 
\begin{equation}
(\Gamma^{(i)} h)(x, s, c)  = (\Gamma^{(i)} f)(x + S_s S_t x_1 - S_t x_2, st, c)
\end{equation}
of the scale channel feature maps
under a more general combined scaling transformation and translation of the form
\begin{equation}
\label{eq-sc-transf-app-sc-cov-proof-noncent-offset}
h(x') = f(x) \quad \mbox{for} \quad x' = S_t (x - x_1) + x_2
\end{equation}
corresponding to
\begin{equation}
h(x) = f(S_t^{-1}(x - x_2) + x_1)
\label{eq:general_scaling_transformation}
\end{equation}
using an integral representation of the deep network.
In the special case when $x_1 = x_2 = x_0$, this corresponds to a
uniform scaling transformation around $x_0$ (i.e. $S_ {x_0,s}$). 
With $x_1 = x_0$ and $x_2 = x_0 + \delta$, this
corresponds to a scaling transformation around $x_0$ followed by a
translation $\calD_\delta$. 

Consider a deep network $\phi^{(i)}$ 
 (\ref{eq:phi_i-def}) 
and assume the integral representation (\ref{eq:phi_integral}), where we for simplicity of notation incorporate the offsets $b_{i,c}$ into the 
non-linearities $\sigma_{i,c}$. 
By expanding the integral representation 
%(\ref{eq:phi_integral}) 
of the rescaled image $h$ (\ref{eq:general_scaling_transformation}),
we have that that the feature representation in the scale channel network is given by
 (with $M_0 = 1$ for a scalar input image):
 
 \begin{align}
 \begin{split}
 &  (\Gamma^{(i)} h)(x, s, c)  =  \{ \mbox{definition (\ref{eq:gamma_s-def})}  \}  
 = (\phi_s^{(i)} h)(x, c) 
 \end{split}\nonumber\\
 \begin{split}
 & = \{ \mbox{definition (\ref{eq:phi_s-def})} \} 
  = (\phi^{(i)} \, h_s)(x, c) 
 = \{ \mbox{equation~(\ref{eq:phi_i-def})} \}
 \end{split}\nonumber\\
 \begin{split}
 & = (\theta^{(i)} \theta^{(i-1)} \dots \theta^{(2)} \theta^{(1)}  h_s)(x, c) 
 = \{ \mbox{equation~(\ref{eq:phi_integral})} \}
 \end{split}\nonumber\\
 \begin{split}
 & = 
 \sigma_{i,c}
 \left(
 \sum_{m_i=1}^{M_{i-1}}
 \int_{\xi_i \in \bbbr^N} 
 \sigma_{i-1,m_i}
 \left(
 \sum_{m_{i-1}=1}^{M_{i-2}}
 \int_{\xi_{i-1} \in \bbbr^N} 
 \dots
 \right.
 \right.
 \end{split}\nonumber\\
 \begin{split}
 & \phantom{\sigma_i \vphantom{\left( \sum_{m_i=1}^{M_{i-1}} \right.)}} \quad
 \left.
 \left.
 \sigma_{1,m_2}
 \left(
 \sum_{m_1=1}^{M_0}
 \int_{\xi_1 \in \bbbr^N} 
 h_s(x - \xi_i - \xi_{i-1} -
 \dots - \xi_1) \, \times
 \right.
 \right.
 \right.
 \end{split}\nonumber\\
 \begin{split}
 & \phantom{\sigma_i \vphantom{\left( \sum_{m_i=1}^{M_{i-1}} \right.)}} \quad
 \left.
 \left.
 \left.
 \phantom{\left(\sum_{m_i=1}^{M_{i-1}} \right.)}
 g_{m_1,m_2}^{(1)}(\xi_1)  \, d\xi_1
 \right) \dots
 g_{m_{i-1},m_i}^{(i-1)}(\xi_{i-1})  \, d\xi_{i-1}
 \vphantom{\left( \sum_{m_i=1}^{M_{i-1}} \right.)} 
 \right)
 \right.
 \end{split}\nonumber\\
 \begin{split}
 \label{eq-exp-int-repr-sc-cov-proof-noncent-offset}
 \left.
 \vphantom{\left(\sum_{m_i=1}^{M_{i-1}} \right.)}\quad\quad
 g_{m_i,c}^{(i)}(\xi_i)  \, d\xi_i
 \vphantom{\left( \sum_{m_i=1}^{M_{i-1}} \right.)} \right).
 \end{split} 
 \end{align}
Under the scaling transformation (\ref{eq-sc-transf-app-sc-cov-proof-noncent-offset}), the part of the integrand 
$h_s(x - \xi_i - \xi_{i-1} - \dots -\xi_1)$ transforms as follows:
\begin{align}
\begin{split}
& h_s(x - \xi_i - \xi_{i-1} - \dots - \xi_1)
\end{split}\nonumber\\
\begin{split}
& = \{ \mbox{$h_s(x) = h(S_s^{-1}x)$ according to definition (\ref{eq:scale-def2})}  \}
\end{split}\nonumber\\
\begin{split}
& = h(S_s^{-1} (x - \xi_i - \xi_{i-1} - \dots - \xi_1))
\end{split}\nonumber\\
\begin{split}
& = \{ \mbox{$h(x) = f(S_t^{-1} (x-x_2) + x_1)$ according to (\ref{eq:general_scaling_transformation}) } \}
\end{split}\nonumber\\
\begin{split}
& = f(S_t^{-1} S_s^{-1} ((x - \xi_i - \xi_{i-1} - \dots - \xi_1) - S_s x_2 + S_s S_t x_1)
\end{split}\nonumber\\
\begin{split}
& = \{ \mbox{$S_s S_t = S_{st}$ for scaling transformations} \}
\end{split}\nonumber\\
\begin{split}
& = f(S_{st}^{-1} ((x + S_s S_t x_1 - S_s x_2 - \xi_i - \xi_{i-1} - \dots - \xi_1))
\end{split}\nonumber\\
\begin{split}
& = \{ \mbox{$f_{st}(x) = f(S_{st}^{-1} x)$ according to definition (\ref{eq:scale-def2})} \}
\end{split}\nonumber\\
\begin{split}
& = f_{st}(x + S_s S_t x_1 - S_s x_2 - \xi_i - \xi_{i-1} - \dots - \xi_1).
\end{split}
\end{align}
Inserting this transformed integrand into the integral representation (\ref{eq-exp-int-repr-sc-cov-proof-noncent-offset})
gives
\begin{align}
\begin{split}
&  (\Gamma^{(i)} h)(x, s, c) =  
\end{split}\nonumber\\
\begin{split}
& = 
\sigma_{i,c} 
\left(
\sum_{m_i=1}^{M_{i-1}}
\int_{\xi_i \in \bbbr^N} 
\sigma_{i-1,m_i}
\left(
\sum_{m_{i-1}=1}^{M_{i-2}}
\int_{\xi_{i-1} \in \bbbr^N} 
\dots
\right.
\right.
\end{split}\nonumber\\
\begin{split}
& \phantom{\sigma_i \vphantom{\left( \sum_{m_i=1}^{M_{i-1}} \right.)}} \quad
\left.
\left.
\sigma_{1,m_2}
\left(
\sum_{m_1=1}^{M_0}
\int_{\xi_1 \in \bbbr^N} 
f_{st}(x + S_s S_t x_1 - S_s x_2  -
\right.
\right.
\right.
\end{split}\nonumber\\
\begin{split}
& \hphantom{\sigma_i \vphantom{\left( \sum_{m_i=1}^{M_{i-1}} \right.)}} \quad
\left.
\left.
\hphantom{\sigma_{1,m_2} \left(  \sum_{m_1=1}^{M_0}
	\int_{\xi_1 \in \bbbr^N} \right. } \quad\quad
\xi_i - \xi_{i-1} - \dots - \xi_1) \times
\right.
\right.
\end{split}\nonumber\\
\begin{split}
& \phantom{\sigma_i \vphantom{\left( \sum_{m_i=1}^{M_{i-1}} \right.)}} \quad
\left.
\left.
\left.
\phantom{\left(\sum_{m_i=1}^{M_{i-1}} \right.)}
g_{m_1,m_2}^{(1)}(\xi_1)  \, d\xi_1
\right) \dots
g_{m_{i-1},m_i}^{(i-1)}(\xi_{i-1})  \, d\xi_{i-1}
\vphantom{\left( \sum_{m_i=1}^{M_{i-1}} \right.)} 
\right)
\right.
\end{split}\nonumber\\
\begin{split}
\left.
\vphantom{\left(\sum_{m_i=1}^{M_{i-1}} \right.)}\quad\quad
g_{m_i,c}^{(i)}(\xi_i)  \, d\xi_i
\vphantom{\left( \sum_{m_i=1}^{M_{i-1}} \right.)} \right),
\end{split} 
\end{align}
which we recognize as
\begin{align}
\begin{split}
&  (\Gamma^{(i)} h)(x, s, c) 
\end{split}\nonumber\\
\begin{split}
& = (\theta^{(i)} \theta^{(i-1)} \dots \theta^{(2)} \theta^{(1)} f_{st})(x + S_s S_t x_1 - S_s x_2, c) 
\end{split}\nonumber\\
\begin{split}
& = (\phi^{(i)} \, f_{st})(x + S_s S_t x_1 - S_s x_2, c) 
\end{split}\nonumber\\
\begin{split}
&  = (\phi_{st}^{(i)} f)(x + S_s S_t x_1 - S_s x_2, c) 
\end{split}\nonumber\\
\begin{split}
&=  (\Gamma^{(i)} f)(x + S_s S_t x_1 - S_s x_2, st, c)
\end{split}
\end{align}
and which proves the result. Note that for a pure translation ($S_t = I$,  $x_1 = x_0 $  and $x_2 = x_0 + \delta$) this gives
\begin{align}
&(\Gamma^{(i)}\, \calD_\delta\, f)(x,c,s) = (\Gamma^{(i)} f)(x - S_s \delta, s, c).
\label{eq:translation-covariance-scale}
\end{align}
Thus, translation covariance is preserved in the scale channel network but the magnitude of the spatial shift in the feature maps will depend on the scale channel.

%the focus is less on that the different scale channels gradually simplifying the image and more on.  

\subsection{Relations to scale-space theory}
In classical scale-space theory \cite{Iij62-TR,Wit83,Koe84-BC,lindeberg1993-scspbook,lindeberg1994-scsparticle,Flo97-book,scalespaceJapan_weickert99,haarromeny-04book}, a multi-scale representation of an input image is created by convolving the image with a set of rescaled and normalised Gaussian kernels. The  scale channel networks described in this paper are  based on a similar philosophy of processing an image at \emph{all scales simultaneously}, although \emph{the input image}, as opposed to the filter, is expanded over scales.
%, is expanded over scales. expanded o instead of the filters are expanded over scales. 
For continuous image data, a representation computed by applying a fixed size filter to a set of rescaled input images is computationally equivalent to applying a set of rescaled and scale-normalised filters to a fixed size input (as done when computing a Gaussian scale-space representation). The two representations are related through a \emph{spatial rescaling} and an \emph{inverse mapping of the scale parameter} $s \mapsto s^{-1}$ (see Appendix \ref{app:relation-to-scale-space1}).
For discrete image data, a similar relation holds approximately, provided that the discrete rescaling operation is a sufficiently good approximation of the continuous rescaling operation. 
% !--> For this to be true we _need_ normalised kernels and we do not have that.

A key difference compared to classical scale-space representations is that \emph{non-linear} feature extractors \emph{learned from data} are used as opposed to the mathematically derived Gaussian derivatives and differential invariants.
%Further, the Gaussian scale space representation is explicitly focused on gradually simplifying the image by suppressing fine scale structures the features extracted in the network can correspond to abritrary non-linear operations learned from data.  
%A scale-channel network could possibly learn to perform a simplification of image structures in some feature channels, but this is not explicitly enforced.
The outputs from the scale channels do, however, still constitute a \emph{(non-linear) scale-covariant multi-scale representation}, which implies that e.g. maxima over scale are preserved, although shifted to a different scale channel, when an input image is rescaled.
% that the classification accuracy metric will encourage kernels that takes a relatively distinct maxima over scales. 
% Thus, e.g. not necessarily fulfilling the property of non-enhancement of local extrema. 

%The use of \emph{max-pooling} over the outputs of the scale channels ($\Lambda_{\max}, \Lambda_{sw,\max}$) is structurally similar to classical methods for \emph{scale selection}, where maxima over scale are detected of scale-normalised filter responses \cite{Lin97-IJCV,Lin14-EncCompVis} or from the scales at which a supervised classifier delivers class labels with the highest posterior \cite{Lin98-IJCV,LiTaxLoo11-ScSp, LooLiTax09-LNCS}. A difference is that, here, max pooling is done over more complex feature responses already adapted to detect specific objects while classical scale selection is performed in a class-agnostic way based on low-level features. 
The use of \emph{supremum}, or for a discrete set of scale channels, \emph{max-pooling}, (see further Section \ref{sec:discrete-scale-channels}) over the outputs of the scale channels is structurally similar to classical methods for \emph{scale selection}, which detect maxima over scale of scale-normalised filter responses \cite{lindeberg97-IJCV,scaleselection-Lin14}. Here, max pooling is, however, done over more complex feature responses, already adapted to detect specific objects, while classical scale selection is performed in a class-agnostic way based on low-level features. This makes max-pooling in the scale channel networks also closely related to more specialized classical methods that detect maxima
 from the scales at which a supervised classifier delivers class labels with the highest posterior \cite{LiTaxLoo11-ScSp, LooLiTax09-LNCS}. 
%This makes the methodology similar to choosing the scale for which the class posterior takes a maximum (ref?).
Average pooling over the outputs of a discrete set of scale channels (Section \ref{sec:discrete-scale-channels}) is structurally similar to methods for scale selection that are based on \emph{weighted averages} of filter responses at different scales \cite{Lin15-JMIV,scaleselection-Lin13JMIV}. Although there is no guarantee that the learned non-linear features will, indeed, take maxima for relevant scales, one might expect training to promote this, since a failure to do so should be detrimental to the classification performance of these networks. 
In case the learned features correspond to partial \emph{Gaussian
derivatives} of some orders, then the
application of these filters to all the scale channels is, in fact, computationally equivalent to applying corresponding
	\emph{scale-normalised Gaussian derivatives} to the original image (see Appendix \ref{app:relation-to-scale-space2}).
%The local extrema over scales of such scale-normalized derivatives reflect characteristic scales in the image data for differential expressions adapted to the visual task. 

%the local extrema over scales of such scale-normalized derivatives are guaranteed to follow scale variations in the input and reflect characteristic scales in the image data provided that the corresponding differential expressions are sufficiently suitable for the visual task.

%provided that the corresponding differential expressions are adapted to the visual task.
% If a scale channel network learns filters that correspond to single scale Gaussian derivatives, the representation in the network is, in fact, equivalent to a representation computed by applying a \emph{scale-normalised} derivatives  %\cite{lindeberg97-IJCV,scaleselection-Lin14} 
%to an input image and these are guaranteed to take maxima for scales corresponding to relevant physical scales in the image data. 
%%In a context of classification there are also large similarities with methods of choosing the class label which has the maximum posterior density. 

\section{Discrete scale channel networks}
\label{sec:discrete-scale-channels}
Discrete scale channel networks are implemented by using a standard discrete CNN as the base network $\phi$. For practical applications, it is also necessary to restrict the network to include a finite number of scale channels $\hat{S} = \{ \gamma^{i}\}_{-K_{min} \leq i \leq K_{max} }$. The input image $f:\bbbz^2 \to \bbbr$ is assumed to be of finite support. 
The outputs from the scale channels are, here,  aggregated using e.g. max pooling
\begin{equation}
(\Lambda_{\max}f)(x,c) = \max_{s \in \hat{S}} \{(\phi_s f)(x,c,s) \} 
\label{eq:lambda-s-max}
\end{equation}
or average pooling
\begin{equation}
(\Lambda_{\avg} f)(x,c)= \mathop{\avg}_{s \in \hat{S}} \{(\phi_s f)(x,c,s) \}.
\label{eq:lambda-s-avg}
\end{equation}
We will also implement discrete scale channel networks that concatenate the outputs from the scale channels followed by an additional transformation $\varphi: \bbbr^{M_i |\hat{S}|} \to \bbbr^{M_i}$ that mixes the information from the different channels
\begin{align} 
&(\Lambda_{\conc} f)(x, c) \nonumber \\
&=  \varphi \left( [(\phi_{s_1} f)(x,c), (\phi_{s_2} f)(x,c) \cdots  (\phi_{s_{|\hat{S}|}} f)(x,c) ] \right).
\label{eq:lambda-s-conc}
\end{align}
$\Lambda_{\conc}$ does not have any theoretical guarantees of invariance, but 
since scale concatenation of outputs from the scale channels has been previously used with the explicit aim of scale invariant recognition \cite{XuXiaZhaYan-arXiv2014}, we will evaluate it also here.

\subsection{Foveated processing}
\label{sec:foveated-operations}
A standard convolutional neural network $\phi$ has a finite support region $\Omega$ in the input. 
When rescaling an input image of fixed size/finite support in the scale channels, it is necessary to decide how to process the resulting images of varying size using a feature extractor with fixed support. One option is to process regions of \emph{constant size} in the scale channels corresponding to regions of \emph{different sizes} in the input image.
This results in \emph{foveated image operations}, where a smaller region around the center of the input image is processed with high resolution, while gradually larger regions of the input image are processed with gradually reduced resolution (see Figure~\ref{fig:foveated-processing}b-c). 
We will refer to the foveated network architectures $\Lambda_{\max} $, $\Lambda_{\avg} $ and $\Lambda_{\conc} $ as the FovMax network, the FovAvg network and the FovConc network respectively.

\subsection{Approximation of scale invariance}
Foveated processing combined with max or average pooling 
will give an approximation of the scale invariance in the continuous model (Section \ref{sec:scale-invariance}) over \emph{a limited scale range}. The numerical scale warpings of the input images in the scale channels approximate continuous scaling transformations. A discrete set of scale channels will approximate the representation for a continuous scale parameter. 
% where the approximation will be better with denser sampling of the scaling group. 
A possible issue is problems at the scale boundaries of a finite scale interval. 
Boundary effects can, however, be mitigated if the network learns to suppress responses for both very zoomed in and very zoomed out objects. % so that these are close to zero. 
If including a large enough number of scale channels and training the network from scratch, this is, in fact, a likely scenario, since the network will otherwise classify based on use of object views that will hardly provide useful information. 

\subsection{Sliding window processing in the scale channels}
\label{sec:sliding-window}

An alternative option for dealing with varying image sizes is to, in each scale channel, process the entire rescaled image by applying the base network in \emph{a sliding window manner}. 
The output from the scale channels can then be combined by max (or average) pooling over space followed by max (or average) pooling over scales
\begin{equation}
(\Lambda_{sw,\max} f)(c) = \max_{s \in S} \max_{x \in \Omega_s} \{ (\phi_s f)(x, c, s)  \},
\label{eq:disc-sliding-window-pool}
\end{equation}
where $\Omega_s = \{sx | x \in \Omega \}$.
We will here only evaluate the architecture using max pooling, which is structurally similar to the popular multi-scale OverFeat detector \cite{SerEigZha-arXiv2013}. This network will be referred to as the SWMax network. For this scale channel network to support invariance, it is not enough that boundary effects resulting from using a finite number of scale channels are mitigated. When processing regions in the scale channels corresponding to only a single region in the input image, new structures can appear (or disappear) in this region for a rescaled version of the original image. With a linear approach this might be expected to not cause problems.
%\footnote{
%	When using linear template matching, the best matching pattern for a template learned during training will be a very similar image patch. Thus, when sliding a template across a matching object it will take the maximum response when \emph{centered} on the object. When using a non-linear method, however, there is no reason there could not be large responses for non centered views of familiar objects or completely novel patterns.}
% since the best matching pattern will be the one corresponding to the template learned during training. 
For a deep neural network, however, there is no guarantee that there cannot be strong erroneous responses for e.g. a partial view of a zoomed in object. We are, here, interested in studying the effects this has on generalisation in the deep learning context. 
%This method does not support invariance to rescalings of an input image even when disregarding boundary effects resulting from using a finite number of scale channels.

\begin{figure*}[h]
	\centering
	\subfloat[Subfigure 1 list of figures text][Standard CNN.]{
		\includegraphics[width=0.48\textwidth]{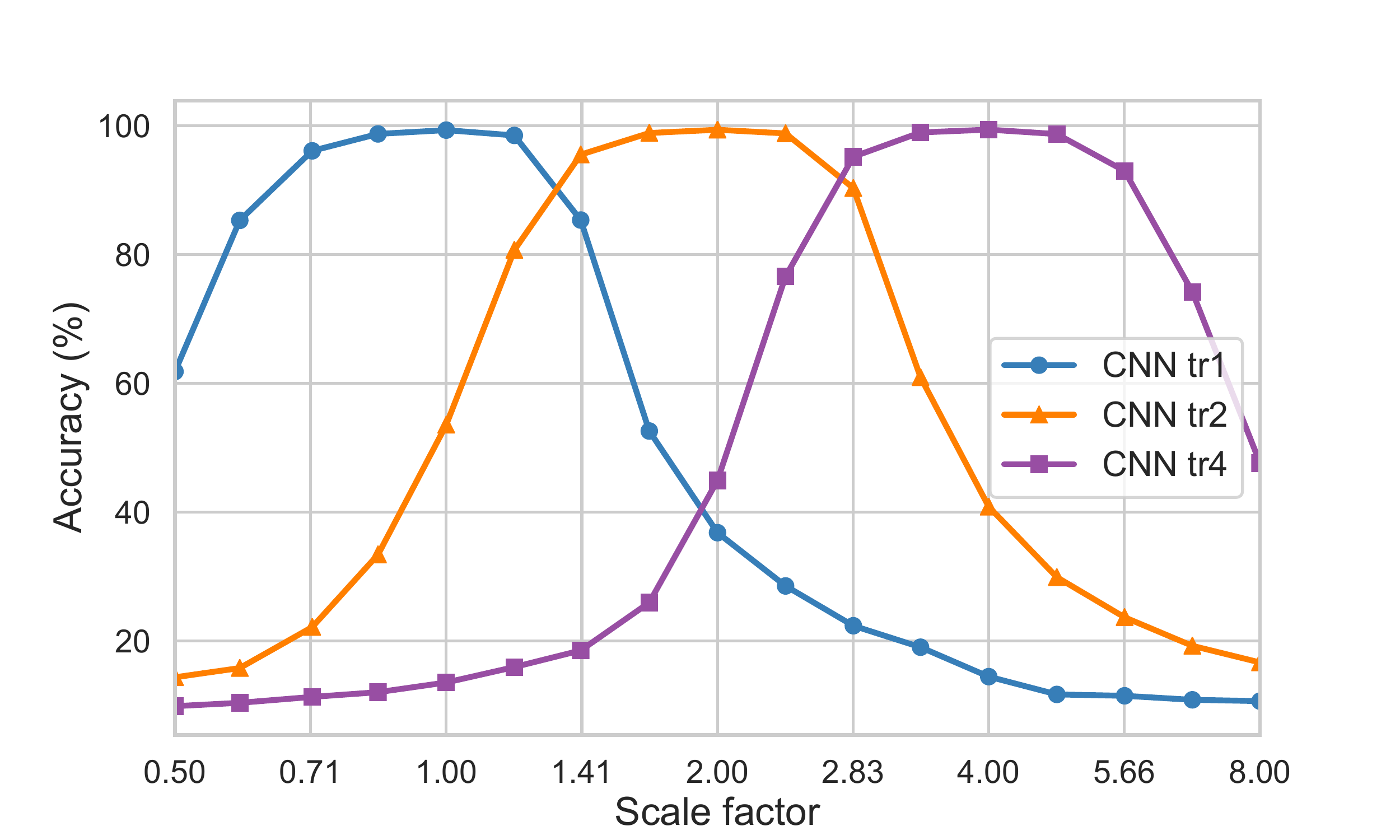}
		\label{fig:subfig1}
}
%	\qquad
	\subfloat[Subfigure 2 list of figures text][The FovConc network.]{
		\includegraphics[width=0.48\textwidth]{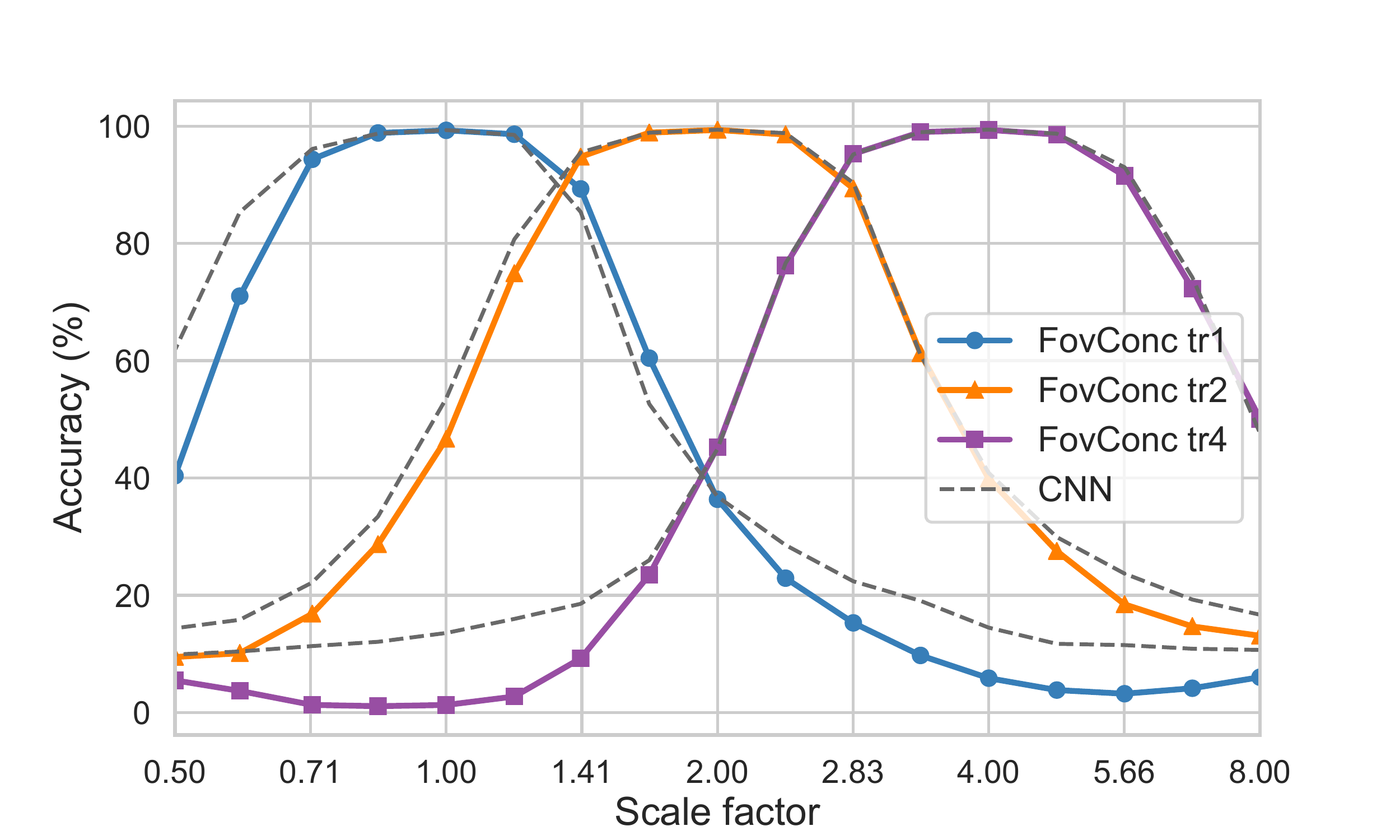}
		\label{fig:subfig2}
} \\
\vspace*{-1em}
	\subfloat[Subfigure 3 list of figures text][The FovMax and FovAvg networks]{
		\includegraphics[width=0.48\textwidth]{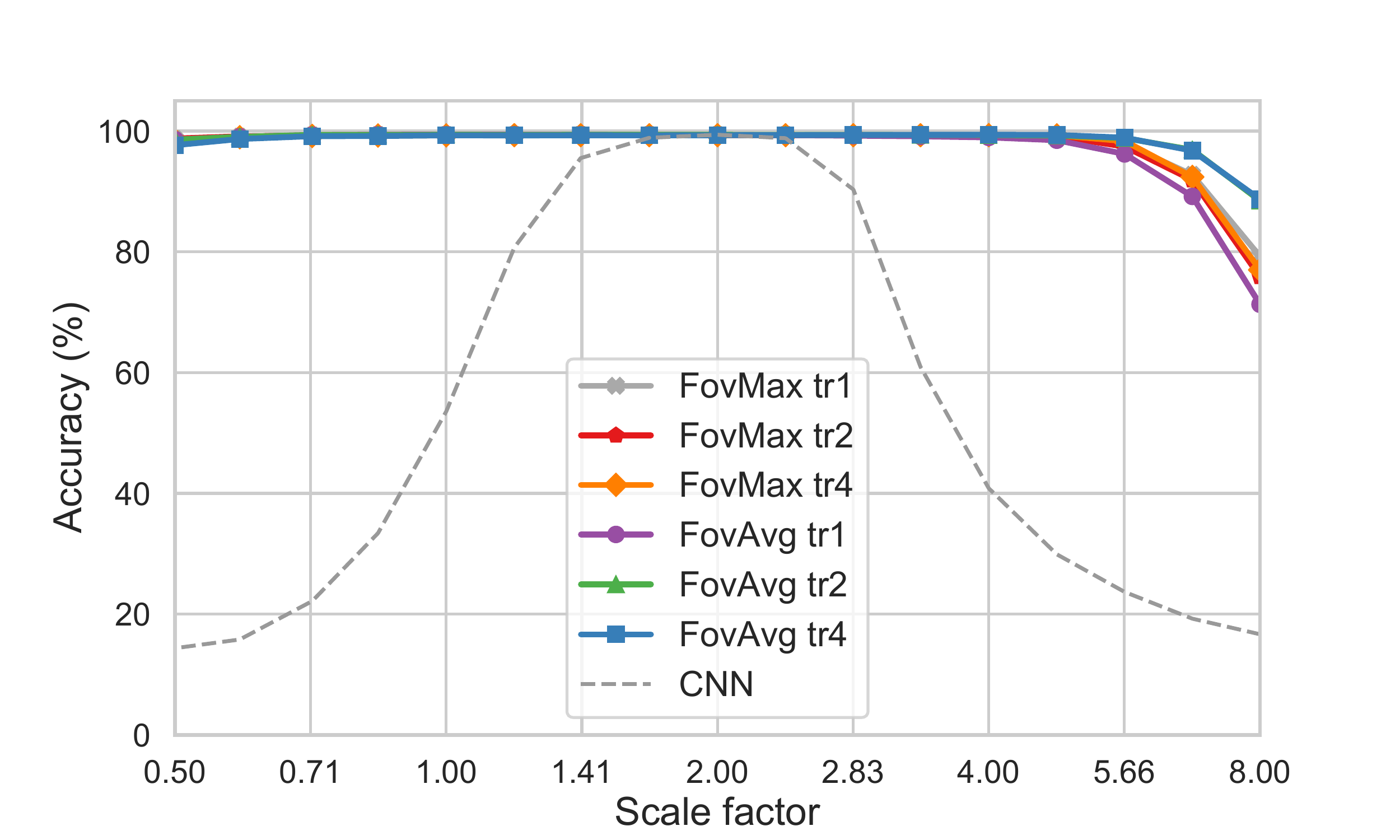}
		\label{fig:subfig3}
}
%	\qquad
	\subfloat[Subfigure 4 list of figures text][The SWMax network]{
		\includegraphics[width=0.48\textwidth]{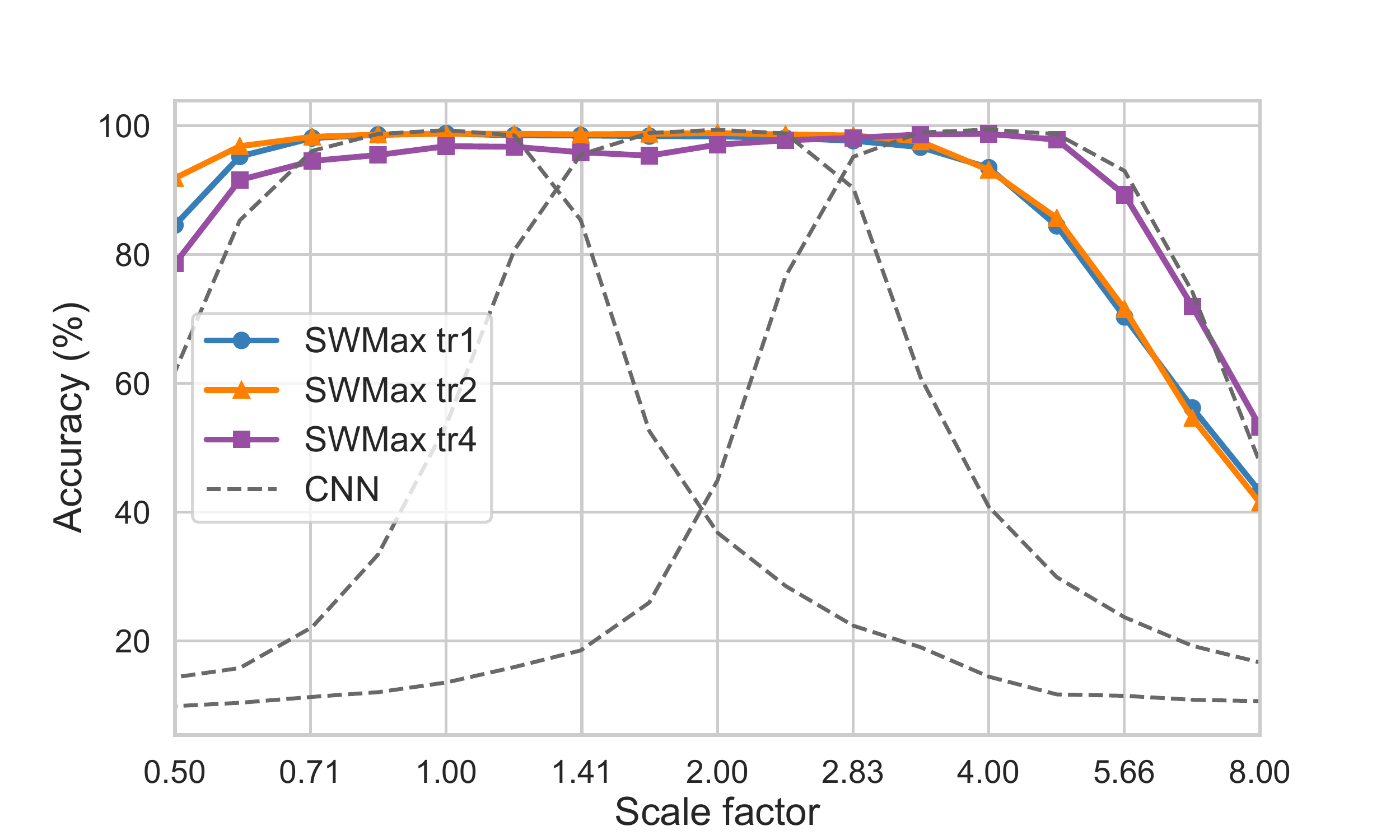}
		\label{fig:subfig4}
}
	\caption{\emph{Generalisation ability to unseen scales for a standard CNN and the different scale channel network architectures}. The networks are trained on digits of scale 1 (tr1), scale 2 (tr2) or scale 4 (tr4) and evaluated for varying rescalings of the test set. We note that the CNN (a) and the FovConc network (b) have poor generalisation ability to unseen scales, while the FovMax and FovAvg networks (c) generalise extremely well. The SWMax network (d) generalises considerably better than a standard CNN, but there is some drop in performance for scales not seen during training. 
	}
	\label{fig:single-scale-generalisation}
\end{figure*}

\section{Experiments}

\subsection{The MNIST Large Scale dataset}
\label{sec:mnist-scale}
To evaluate the ability of standard CNNs and  scale channel networks to generalise to unseen scales over \emph{a wide scale range}, we have created a new version of the standard MNIST dataset \cite{MNISToriginal-1998}. 
This new dataset, \emph{MNIST Large Scale}, which is available online \cite{MnistLargeScale-online}, is composed of images of size $112 \times 112$  with scale variations of a factor 16 for scale factors $s \in [0.5, 8]$ relative to the original MNIST dataset. The train and test sets for the different scale factors are created by resampling the original MNIST training and test sets using bicubic interpolation followed by smoothing and soft thresholding to reduce discretization effects. Note that for scale factors $>4$, the full digit might not be visible in the image. These scale values are nonetheless included to study the limits of generalisation. More details concerning this dataset are given in Appendix \ref{app:mnist-large-scale}.

\subsection{Network and training details}

\emph{The standard CNN} is composed of 8 conv-batchnorm-ReLU blocks followed by a fully connected layer and a final softmax layer. The number of features/filters in each layer is 16-16-16-16-32-32-32-32-100-10. A stride of 2 is used in convolutional layers 2, 4, 6 and 8. The reason for using a quite deep network is to avoid a network structure that is heavily biased towards recognising either small or large digits.

\emph{The FovMax, FovAvg, FovConc and SWMax\footnote{We noted that batchnorm impairs performance when training the SWMax network from scratch. We believe this is because the sliding window approach implies in a change
	in the feature distribution when processing data of different scales.
	% For batchnorm to function optimally the data/feature distribution should stay approximately the same. 
	We, therefore, train the SWMax network without batchnorm.} 
scale channel networks} are constructed using scale channels with 4 conv-batchnorm-ReLU blocks followed by a fully connected layer and a final softmax layer. Rescaling within the scale channels is done with bilinear interpolation and applying border padding or cropping as needed. Batchnorm layers are shared across the scale channels. 
The number of features/filters in each layer is 16-16-32-32-100-10. A stride of 2 is used in convolutional layers 2 and 4. All scale channel architectures have around 70\,000 parameters, while the baseline CNN has around 90\,000 parameters.

All networks are trained with 50\,000 training samples from the MNIST Large Scale dataset for 20 epochs using the Adam optimiser. During training, 15 \% dropout is applied to the first fully connected layer. The learning rate starts at $3e^{-3}$ and decays with a factor $1/e$ every second epoch towards a minimum learning rate of $5e^{-5}$. Results are reported for the MNIST Large Scale test set (10\,000 samples) as the average of training each network using three different random seeds. The remaining 10\,000 samples constitute a validation set. Numerical performance scores for Figures~2--5 are given in Appendix \ref{app:numerical}.

\subsection{Generalisation to unseen scales}
We, first, evaluate the ability of the standard CNN and the different scale channel networks to generalise to previously unseen scales. We train each network on each of the scales 1, 2, and 4 
%from the MNIST Large Scale dataset 
and evaluate the performance on the test set for scale factors between $1/2$ and $8$. The FovMax, FovAvg and SWMax networks have 17 scale channels spanning the scale range $[\frac{1}{2}, 8]$. The FovConc network has 3 scale channels spanning the scale range $[1,4]$.\footnote{The FovConc network performs considerably worse when including too many scale channels or spanning a too large scale range. Since we are more interested in the best case rather than the worst case scenario, we, here, picked the best network out of a large range of configurations.} The results are presented in Figure \ref{fig:single-scale-generalisation}. 
We, first, note that all networks achieve similar top performance for the scales seen during training. There are, however, large differences in the abilities of the networks to generalise to unseen scales: 

\subsubsection{Standard CNN}
The standard CNN shows limited generalisation ability to unseen scales with a large drop in accuracy for scale variations larger than a factor $\sqrt{2}$. This illustrates that, while the network can recognise digits of all sizes, a vanilla CNN includes no structural prior to promote scale invariance. 

\subsubsection{The FovConc network}
 The generalisation ability of the FovConc network is quite similar to that of the standard CNN, sometimes slightly worse. The reason for limited generalisation is that although the scale channels share weights, when simply concatenating the outputs from the scale channels there is no structural constraint to support invariance. This is consistent with our observation that 
  spanning a too large scale range or using too many channels degrades generalisation for the FovConc network. For scales \emph{not present during training}, there is, simply, no useful training signal to learn the correct weights in the fully connected layers combining the scale channel outputs. 
  Note that our results are not contradictory to those previously reported for a similar network structure  \cite{XuXiaZhaYan-arXiv2014}, since they train on data that contain natural scale variations and test over a quite narrow scale range. What we do show, however, is that this network structure is \emph{not scale invariant}.

\begin{figure}[hbpt]
	\begin{center}
		\includegraphics[width=0.5\textwidth]{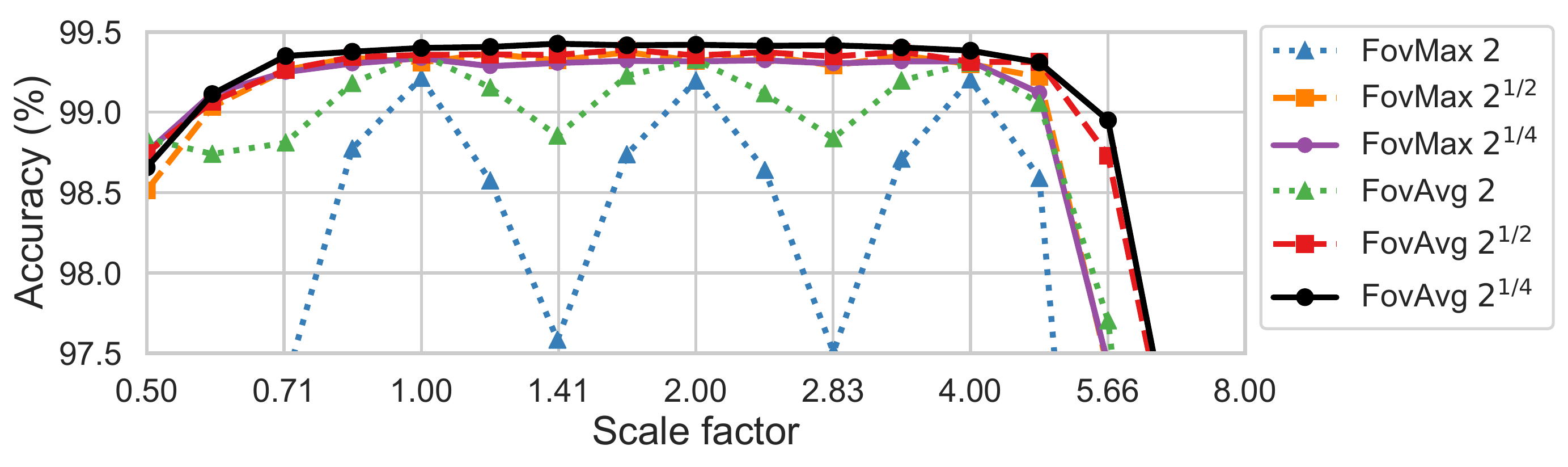}   
	\end{center}
	\caption{\emph{Varying the sampling density of the scale channels.} FovMax and FovAvg networks spanning the scale range $[\frac{1}{4},8]$ are trained with varying spacing between the scale channels ($2$, $2^{1/2}$ and $2^{1/4}$). All networks are trained on scale 2. There is a significant increase in the performance when reducing the spacing between the scale channels from $2$ to $2^{1/2}$ while the effect of a further reduction to $2^{1/4}$ is small. 
		%from the MNIST Large Scale dataset. 
	}
	\label{fig:denseness}	
\end{figure}

\subsubsection{The FovAvg and FovMax networks}
We note that the FovMax and FovAvg networks generalise very well, independently of which scale the network is trained on. The maximum difference in performance in the scale range $[1,4]$ between training on scale 1, scale 2 or scale 4  is less than 0.2 percentage points for these network architectures.
Importantly, this shows that, if including a large enough number of scale channels and training the networks from scratch, boundary effects at the scale boundaries do not prohibit invariant recognition. 
For the FovAvg and FovMax networks, we also investigate how densely it is necessary to sample the scale channels for good performance. The result is presented in Figure \ref{fig:denseness}. Accuracy is considerably improved when decreasing the distance between consecutive channels from a factor $2$ (5 channels) to a factor of $2^{1/2}$ (9 channels), while a further reduction to $2^{1/4}$ (17 channels) provide very small additional benefits.

\begin{figure*}[hbpt]

\begin{tabular}{cc}
			\includegraphics[width=0.48\textwidth]{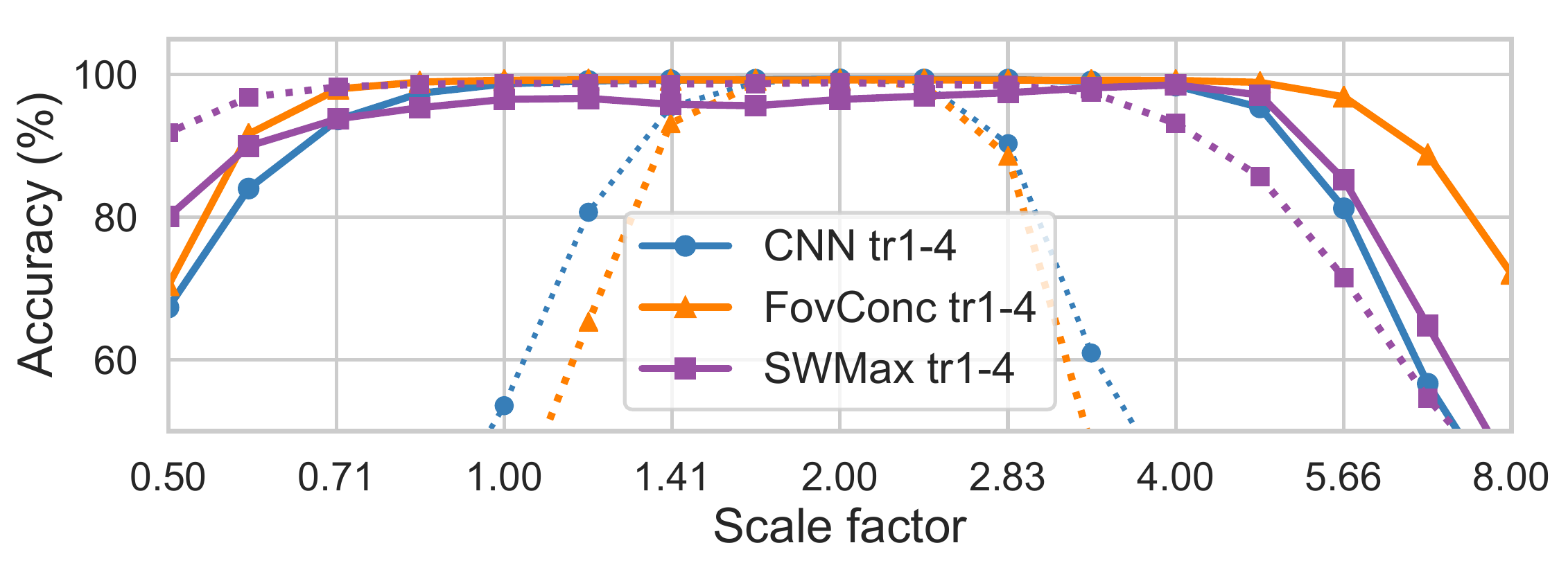}  & 
				\includegraphics[width=0.48\textwidth]{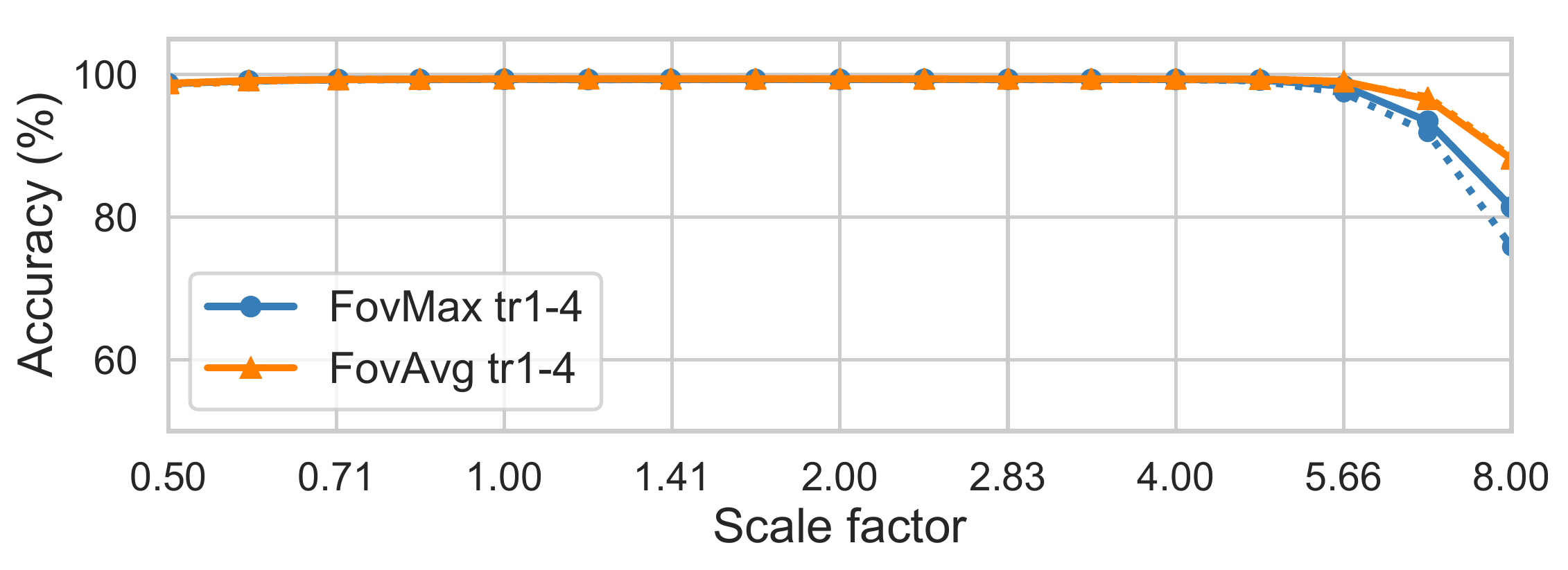} 
\end{tabular}

	\caption{\emph{Multiscale image data.}
		All networks are trained on digits in the scale range $[1,4]$  (tr1-4) and evaluated for varying scale factors in the test set. 
		The difference in generalisation ability between training on multiscale and single scale data (dotted lines) is striking for both the CNN and the FovConc network. %For the SWMax network, including multiscale data improves generalisation somewhat for larger scales but impairs generalisation somewhat for smaller scales. 
		For the FovMax and FovAvg networks, the difference is negligible between multiscale and single scale training, which illustrates the strong invariance properties of these networks.}
	\label{fig:compare_multiscale_all}	
\end{figure*}

\subsubsection{The SWMax network}
We note that the SWMax network generalises considerably better than a standard CNN, but there is some drop in performance for scales not seen during training. 
We believe that the main reason for this is, here, that since all scale channels are processing a fixed sized region in the input image (as opposed to for foveated processing), new structures might leave or enter this region when an image is rescaled. This might lead to erroneous high responses for unfamiliar views (Section \ref{sec:sliding-window}). 
We also noted that the SWMax networks are harder to train (more sensitive to learning rate etc) compared to the foveated network architectures as well as more computationally expensive. 
Thus, while the FovMax and FovAvg networks still are easy to train and the performance is not degraded when spanning a large scale range, the SWMax network seems to work best for spanning a more limited scale range where fewer scale channels are needed (as was indeed the use case in \cite{SerEigZha-arXiv2013}).

\subsection{Multiscale vs.\ single scale training}
\label{sec:multiscale_vs_singlescale}
All the scale channel architectures support multiscale processing although they might not support scale invariance. We, here, test the performance of the different scale channel networks when training on multiscale training data. 
For the FovMax, FovAvg and FovSW network, the same scale channel setup (17 channels) is used as for single scale training. For the FovConc network, 5 scale channels spanning the scale range $[\frac{1}{2}, 8]$ are used, since this setup gives  better results compared to the previous setup with 3 channels.
% for the single scale training. 
% using the same setup as we used for single scale training (3 channels).
% for this network.

 The results are presented in Figure \ref{fig:compare_multiscale_all}. 
 The difference between training on multiscale and single scale data is striking for the standard CNN and the FovConc network. It can, however, be noted that the FovConc network does generalise slightly better than a standard CNN outside the scale range it is trained on. For the SWMax network, including multiscale data improves generalisation somewhat for larger scales but impairs generalisation somewhat for smaller scales.  
The difference in generalisation ability between training on a single scale or multiscale image data is almost indiscernible for the FovMax and FovAvg networks. 

\subsection{Generalisation from fewer training samples}
\begin{figure}[hbpt]
	
	\begin{center}
		\includegraphics[width=0.45\textwidth]{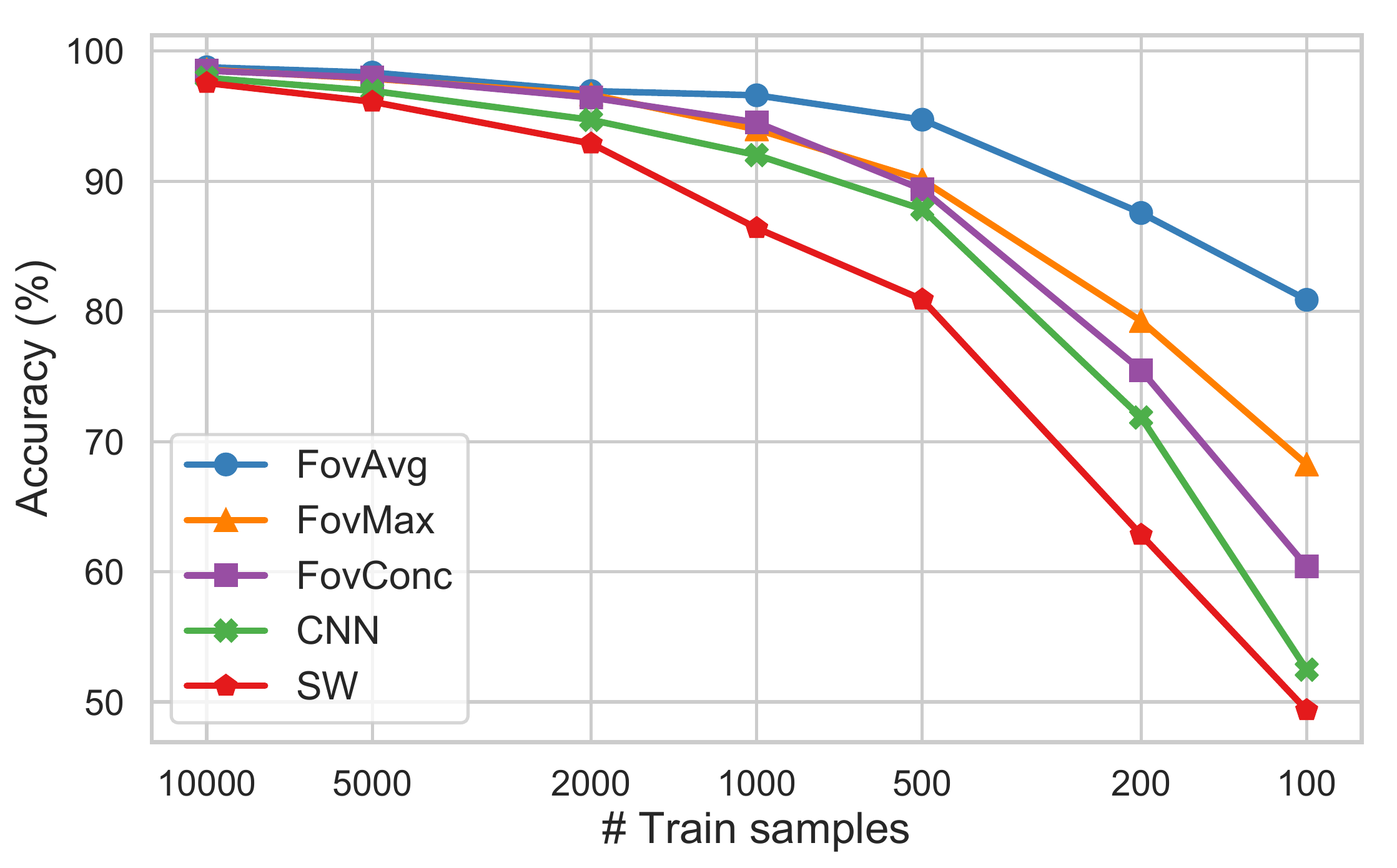}
		
	\end{center}
	\caption{\emph{Training  with smaller training sets with large scale variations}. All network architectures are evaluated on their ability to classify data with large scale variations while reducing the number of training samples. Both the training and test set here span the scale range $[1,4]$. The FovAvg network shows the highest robustness when decreasing the number of training samples followed by the FovMax network. 
   %The FovConc network also shows a small improvement over the standard CNN.
}
	\label{fig:mnist_few_samples}
\end{figure}

Another scenario of interest is when the training data does span a relevant range of scales, but there are few training samples. Theory would predict a correlation between the performance in this scenario and the ability to generalise to unseen scales. 
% scenario for the networks that show stronger ability to generalise to unseen scales (FovAvg, FovMax). 
%Also, all scale channel network architectures might be expected to perform better than a standard CNN because of the weight sharing between the scale channels.
% implies sharing of statistical strength between scales. %)since there are fewer weights and a pattern learnt at a single scale automatically is available at all scales. 
To test this prediction, we trained the standard CNN and the different scale channel networks on multi scale training data spanning the scale range $[1,4]$, while gradually reducing the number of samples in the training set. Here, the same scale channel setup with 17 channels spanning the scale range $[\frac{1}{2}, 8]$ is used for all the architectures. The results are presented in Figure \ref{fig:mnist_few_samples}. We note that the FovConc network shows some improvement over the standard CNN. The SWMax network, on the other hand, does not, and we hypothesise that when using fewer samples, the problem with partial views of objects (see Section \ref{sec:sliding-window}) might be more severe.
 Note that the way the OverFeat detector is used is the original study \cite{SerEigZha-arXiv2013}, is more similar to our single scale training scenario, since they use base networks pretrained on ImageNet.  The FovAvg and FovMax networks show the highest robustness also in this scenario. This illustrates that these networks can give improvements when multiscale training data is available but there are few training samples. 

\section{Summary and conclusions}

We have presented a theoretical analysis of covariance and invariance properties of continuous scale channel networks. Moreover, we performed an experimental evaluation of different types of discrete scale channel networks on the task of generalising to unseen scales over wide scale ranges. The tested networks include a new family of scale channel networks that combine foveated processing with max or average pooling over the scale channels (the FovMax and FovAvg networks). The experimental evaluation illustrates the strong invariance properties of these networks in practice and limitations of previous approaches and vanilla CNNs.
We believe that our proposed foveated scale channel networks
will prove useful in situations where a simple approach that can generalise to unseen scales or learn from small datasets with large scale variations is needed. This type of foveated scale invariant processing could also be included as subparts in more complex frameworks dealing with large scale variations. 

A more overarching aim of this study have been to test the limits of CNNs to generalise to unseen scales over a wide scale range. The key take home message is a proof of concept that such generalisation is possible if including structural assumptions about scale in the network design.

\appendix

\subsection{Relations between scale channel networks and scale-space theory}
\label{app:relation-to-scale-space1}
We, here, discuss in more detail the relationship between the representations computed in a (continuous) scale channel network and the representations computed within classical scale-space theory. Although a multi-layer scale channel network will compute more complex non-linear features, it is enlightening to investigate whether the network could learn to express operations similar to those used within the classical scale-space paradigm. This will increase our confidence that scale channel networks could be expected to work well together with e.g. max-pooling over scales. 

\subsubsection{Preliminaries}
In classical scale-space theory, a \emph{scale-space representation} of an input image $f: \mathbb{R}^N \to \mathbb{R}$ is defined as \cite{lindeberg1993-scspbook}:
\begin{equation}
L(x; \sigma) = \int_{u \in \bbbr^N} f(x - u)\,g(u; \sigma)\,du,
\end{equation}
where $g: \bbbr^N\times\bbbr^+ \to \bbbr$ denotes the (rotationally symmetric) Gaussian kernel
\begin{equation}
g(x;\sigma) =  \frac{1}{(\sqrt{2\pi} \sigma)^{N}}  e^{\frac{-x^2}{2 \sigma^2}},
\end{equation}
and we use $\sigma$ as the the \emph{scale parameter} compared to the more commonly used $t = \sigma^2$.
From this representation, a family of \emph{Gaussian derivatives} can be computed as
\begin{equation}
L_{x^{\alpha}}(x;\sigma) = \partial_{x^\alpha} L(x;\sigma) = ((\partial_{x^{\alpha}} g(\cdot;\; \sigma)) * f(\cdot))(x),
\end{equation}
where $n \in \mathbb{Z}$ and we use multi index notation $\alpha = (\alpha_1, \cdots \alpha_N)$ such that
$\partial_{x^\alpha} = \partial_{x^{\alpha^1}} \cdots \partial_{x^{\alpha^N}}$. The scale-covariance property of the scale-space representation also transfers to such Gaussian derivatives, and 
%\begin{equation}
%L_{x_1^{n} x_2^m} = \partial_{x^n} \d_{y^m} L(\cdot, \cdot, t) = \d_{x^n} \d_{y^m} (g(\cdot, \cdot;t) * f(\cdot, \cdot))
%\end{equation}
%where . 
%The representations computed by convolution with such Gaussian kernels and Gaussian derivatives are scale covariant \cite{lindeberg1993-scspbook}. 
these visual primitives have been widely used within the classical computer vision paradigm to construct scale-covariant and scale-invariant feature detectors and image descriptors \cite{Lin97-IJCV,Lin98-IJCV,BL97-CVIU,ChoVerHalCro00-ECCV,MikSch04-IJCV,Low04-IJCV,BayEssTuyGoo08-CVIU,TuyMik08-Book,Lin13-ImPhys,Lin15-JMIV}. 
%used for designing a large number of scale-covariant and scale-invariant feature detectors and image descriptors in classical computer vision [cite what?].
One way to achieve scale invariance is to first perform \emph{scale selection} and then e.g. extract features at the identified scale. Scale selection can be done by comparing the magnitudes of $\gamma$-normalised derivatives \cite{lindeberg97-IJCV}:
\begin{equation}
\partial_{\xi^{\alpha}} = \partial_{x^\alpha, \gamma-norm} = t^{|\alpha[\gamma/2} \, \partial_{x^{\alpha}} = \sigma^{|\alpha| \gamma} \, \partial_{x^{\alpha}} 
\end{equation}
%\begin{equation}
%\partial_{\xi^{\alpha}} = \partial_{x^\alpha, \gamma-norm} = s^{|\alpha|\gamma/2} \partial_{x^\alpha}
%%\partial \xi = t^{\gamma/2} \partial_{x_1} \text{~~~~} \partial \eta = t^{\gamma/2} \partial_ {x_2}
%\end{equation}
%for the partial derivatives in $N$ dimensions and with $|\alpha| = \alpha_1 + \cdots \alpha_N$
with $\gamma \in [0,1 ]$ as a free parameter and $|\alpha| = \alpha_1 + \cdots + \alpha_N$. Such derivatives are guaranteed to take maxima at scales corresponding to the relevant physical scales of objects in the image. We will here consider the maximally scale-invariant case with $\gamma=1$ 
\begin{equation}
\partial_{\xi^{\alpha}} = \sigma^{|\alpha|}  \partial_{x^\alpha}
\end{equation}
and show that
scale channel networks will 
compute something similar to such scale-normalised derivatives. First, we will, however, consider the relationship between multi-scale representations computed by applying a set of \emph{rescaled kernels} to a single scale image and representations computed by applying the same kernel to a set of \emph{rescaled images}. 

\subsubsection{Scaling the image vs scaling the filter}
Since the scale-space representation can be computed using a single convolutional layer we, here,  compare with a single layer scale-channel network.
We consider the relationship between representations computed by:  
\begin{enumerate}[(i)]	
	\item Applying a set of rescaled and scale-normalised filters (this corresponds to normalising filters to constant $L_1$-norm over scales) $h: \mathbb{R}^N \to \mathbb{R}$ 
	\begin{equation}\label{eq:scaled_and_normalised_filters}
	h_s(x) = \frac{1}{s^{N}} h(\frac{x}{s})
	\end{equation}
	to a fixed size input image $f(x)$: 
	\begin{equation}\label{eq:Lh}
	L_h(x;s) = (f*h_s)(x) = \int_{u \in \bbbr^N} f(u)\,h_s(x - u)\,du,
	\end{equation}
where the subscript indicates that $h$ might not necessarily be a Gaussian kernel. If $h$ is a Gaussian then $L_h = L$. 
	
	\item Applying a fixed size filter $h$ to a set of rescaled input images 
	\begin{equation}
	\label{eq:Mh}
	M_h(x;s) = (f_s*h)(x) = \int_{u \in \bbbr^N} f_s(u)\,h(x - u)\,du,
	\end{equation}
	with
	\begin{equation}
	f_s(x) = f(\frac{x}{s}).
	\end{equation}
	This is the representation computed by a single layer in a (continuous) scale channel network. 
\end{enumerate}
It is straightforward to show that these representations are computationally equivalent and related by a family of scale dependent scaling transformations. %Since we are, here, mainly interested in comparing scale channel networks with the linear representations 
We compute using the change of variables $u =s\,v$, $du = s^{N} dv$:%, $u = sv$
\begin{align}\label{eq:filter-to-image}
L_h(x;s) &=  (f*h_s)(x) \nonumber \\
&= \int_{u \in \bbbr^N} f(x-u)\, \frac{1}{s^{N}}h(\frac{u}{s})\,du \nonumber \\
&= \int_{u \in \bbbr^N} f(x - s v )\, \frac{1}{s^{N}}h(v)\, s^{N}dv \nonumber \\
&= \int_{u \in \bbbr^N} f(s(\frac{x}{s}-v))\, h(v)\,dv \nonumber \\
&= \int_{u \in \bbbr^N} f_{s^{-1}}(\frac{x}{s}-v)\,h(v)\,dv \nonumber \\
&= (f_{s^{-1}} * h)(\frac{x}{s},s^{-1}).
\end{align}
Comparing this with (\ref{eq:Mh}) 
we see that the two representations are related according to 
\begin{equation}\label{eq:l_to_m}
L_h(x;s) = M_h(\frac{x}{s};s^{-1}).
\end{equation}
%An very similar computation (show?) shows that equivalently 
%\begin{equation}\label{eq:m_to_l}
%M_h(x;s) = L_h(\frac{x}{s}, s^{-1}).
%\end{equation}
%Equivalently, using the same change of variables $v=\frac{u}{s}$, $du = s^{D} dv$ 
%\begin{align}\label{eq:image-to-filter}
%M_h(x;s) &= \nonumber \\
%& = (f_s*g)(x) \nonumber \\
%&= \int_{u \in \bbbr^N} f(\frac{u}{s}) g(u-x) du \nonumber \\
%&= \int_{u \in \bbbr^N} f(v)g(sv - x) s^{N} dv \nonumber \\
%&= \int_{u \in \bbbr^N} f(v) s^{N} g(s(v - \frac{x}{s}) dv \nonumber \\
%&= \int_{u \in \bbbr^N} f(v) g_{s^{-1}}(v - \frac{x}{s}) dv \nonumber \\
%&= (f*g_{s^{-1}})(\frac{x}{s}) \nonumber \\
%\end{align}
%it is clear by comparing this result with \ref{eq_}
%$$
%M_h(x;s) = L_h(\frac{x}{s},s^{-1})
%$$
%Thus, e.g. for a certain scale $s$, convolving  an image $f(x)$ with a scaled filter $h_s(x)$ corresponds to convolving an inversely scaled image $f_{s^{-1}}(x)$ with a single scale filter $h(x)$ and then scaling the result back up to the original size. 
We note that the relation (\ref{eq:l_to_m}) preserves \emph{the relative scale} between the filter and the image for each scale and that both representations are scale covariant.
%The reason the spatial transformation ends up as $\frac{x}{s}$ in both cases is because of the inverse relationship between the scale levels in the both representations.
%Both representations are fully scale covariant and can be transformed to each other. 
Thus, to convolve a set of rescaled images with a single scale filter, as done in the scale channel networks, is computationally equivalent to convolving an image with a set of rescaled filters that are $L_1$-normalised over scale. The two representations are related through a \emph{spatial rescaling} and an \emph{inverse mapping of the scale parameter} $s \mapsto s^{-1}$.
Note that it is straightforward to show, using the integral representation of a scale channel network (\ref{eq:phi_integral}), that a corresponding relation between scaling the image and scaling the filters holds for a multi-layer scale channel network as well. 

%A multi-layer scale channel network could indeed learn a kernel corresponding to a Gaussian or Gaussian derivative in deeper layers of the network. 
 % by an equivalent proof for a multi-layer scale channel network using the integral representation (\ref{eq:phi_integral}) that the corresponding holds for a scale channel network with several layers,
%we conclude that the representation computed by applying a single scale filter to a set of rescaled images is equivalent to the representation computed by applying a set of $L_1$-normalised and rescaled filters to a single image. If normalising a filter to constant $L_1$-norm over scales it is equivalent to convolve an image with a set of rescaled filters and to apply a fixed size filter to a set of rescaled images because of the scale covariance property [ref]

The result (\ref{eq:l_to_m}) implies that if a scale channel network learns a feature corresponding to a Gaussian with the standard deviation $\sigma$, %would have the equivalent set of kernels 
then the representation computed by the scale channel network is computationally equivalent to applying the family of kernels
\begin{equation}
h_s(x) = \frac{1}{s^N} h(\frac{x}{s}) = \frac{1}{(\sqrt{2\pi} s\sigma)^{N}}  e^{\frac{-x^2}{2 (s\sigma)^2}}
\end{equation}
 to the original image, given the complementary scaling transformation (\ref{eq:l_to_m}) with its
associated inverse mapping of
the scale parameters $s \mapsto s^{-1}$.
%(with the complementary inverse mapping of the scale parameters $s \mapsto s^{-1}$) (\ref{eq:scaled_and_normalised_filters})
% If $h_x(s) = g(x; sigma^2)$ 
%%with standard deviation $\sigma$
%\begin{equation}
%h(x) = \frac{1}{\sqrt{2\pi}\sigma }  e^{\frac{-x^2}{2 \sigma^2}} 
%\end{equation}
Since this is a family of rescaled and $L_1$-normalised Gaussians, the scale channel network will compute a representation computationally equivalent to a Gaussian scale-space representation. %, where the two representations are related through a spatial rescaling and

\subsection{Relation between scale channel networks and scale-normalised derivatives}
\label{app:relation-to-scale-space2}
We, here, describe the relationship between scale channel networks and scale-normalised derivatives. Assume that a scale channel network in some
layer learns a kernel that corresponds to \emph{a Gaussian
derivative}. We will show that when this kernel
is applied to all the scale channels this correspond to a normalisation over scales of the kernels that is equivalent to applying scale-normalised derivatives at different scales in a scale-space representation of the original image. 
%We will show that when this kernel is applied to all the scale channels, the equivalent set of kernels (\ref{eq:scaled_and_normalised_filters}) will correspond to  \emph{scale-normalised derivatives} evaluated at different scales in a scale-space representation of the original image.

\subsubsection{Preliminaries: Gaussian derivatives in terms of Hermite
	polynomials}

As a preparation for the intended result, we will first establish a
relation between Gaussian derivatives and probabilistic Hermite
polynomials.
The probabilistic Hermite polynomials $H e_n(x)$ are in 1-D defined by the
relationship
\begin{equation}
H e_n(x) = (-1)^n e^{x^2/2} \, \partial_{x^n} \left( e^{-x^2/2} \right)
\end{equation}
implying that
\begin{equation}
\partial_{x^n} \left( e^{-x^2/2} \right) =  (-1)^n H e_n(x) \, e^{-x^2/2} 
\end{equation}
and 
\begin{equation}
\partial_{x^n} \left( e^{-x^2/2\sigma^2} \right) =  (-1)^n H e_n(\frac{x}{\sigma}) \, e^{-x^2/2\sigma^2} \frac{1}{\sigma^n}.
\end{equation}
Applied to a Gaussian function in 1-D, this implies that
\begin{align}
\begin{split}
\partial_{x^n} \left( g(x;\; \sigma) \right) =
\end{split}\nonumber\\
\begin{split}
= \frac{1}{\sqrt{2 \pi} \sigma} \partial_{x^n} \left( e^{-x^2/2\sigma^2} \right) 
\end{split}\nonumber\\
\begin{split}
= \frac{1}{\sqrt{2 \pi} \sigma}  \frac{(-1)^n}{\sigma^n}  H e_n(\frac{x}{\sigma}) \, e^{-x^2/2\sigma^2}
\end{split}\nonumber\\
\begin{split}
\label{eq-gauss-der-herm-pol}
= \frac{(-1)^n}{\sigma^n} H e_n(\frac{x}{\sigma}) \, g(x;\; \sigma).
\end{split}
\end{align}

\subsubsection{Scaling relationship for Gaussian derivative kernels}

Let us assume that the scale channel network at some layer has
learned a kernel
%receptive field 
that corresponds to a Gaussian partial derivative at
some scale $\sigma$:
%here written in terms of multi-index notation in $N$ dimensions with 
%$x = (x_1, x_x, \dots, x_N)$ and $\alpha = (\alpha_1, \alpha_2, \dots, \alpha_N)$:
\begin{align}
\begin{split}
\partial_{x^{\alpha}} g(x;\; \sigma) =
\end{split}\nonumber\\
\begin{split}
= \partial_{x_1^{\alpha_1} x_2^{\alpha_2} \dots x_N^{\alpha_N}}
g(x;\; \sigma) 
%\end{split}\nonumber\\
%\begin{split}
= g_{x_1^{\alpha_1} x_2^{\alpha_2} \dots x_N^{\alpha_N}}(x;\; \sigma)
\end{split}
\end{align}
For later convenience, we write this learned kernel as a scale-normalised
derivative at scale $\sigma$ for $\gamma = 1$ multiplied by some constant $C$:
\begin{equation}
h(x) = C \, \sigma^{\alpha_1 + \alpha_2 + \dots + \alpha_N} g_{x_1^{\alpha_1} x_2^{\alpha_2} \dots x_N^{\alpha_N}}(x;\; \sigma).
\end{equation}
Then, the corresponding family of equivalent kernels $h_s(x)$ in the dual
representation (\ref{eq:Lh}), 
which represents the same effect on the original image as applying the kernel $h(x)$ 
to a set of rescaled images $f_s(x) = f(x/s)$,
% in $N$ dimensions 
provided that a 
complementary scaling transformation and the inverse mapping of the scale parameter $s \mapsto s^{-1}$ are performed,
%is applied to the scale
%parameter, 
is given by
\begin{align}
\begin{split}
h_s(x) 
= \frac{1}{s^N} \, h(\frac{x}{s})
\end{split}\nonumber\\
\begin{split}
= \frac{C}{s^N} \, \sigma^{\alpha_1 + \alpha_2 + \dots + \alpha_N} g_{x_1^{\alpha_1} x_2^{\alpha_2} \dots x_N^{\alpha_N}}(\frac{x}{s};\; \sigma).
\end{split}
\end{align}
Using Eq.~(\ref{eq-gauss-der-herm-pol}) with 
\begin{equation}
g(x;\; \sigma) = \frac{1}{(\sqrt{2 \pi} \sigma)^N} \, e^{-(x_1^2 + x_2^2 + \dots + x_N^2)/2\sigma^2} 
\end{equation}
in $N$ dimensions, we obtain
\begin{align}
\begin{split}
h_s(x) =\frac{C}{s^N} \, \sigma^{\alpha_1 + \alpha_2 + \dots + \alpha_N}
(-1)^{\alpha_1 + \alpha_2 + \dots + \alpha_N}
\end{split}\nonumber\\
\begin{split}
\phantom{=}
He_{\alpha_1}(\frac{x_1}{s\sigma}) \,  He_{\alpha_2}(\frac{x_2}{s\sigma}) \dots He_{\alpha_N}(\frac{x_N}{s\sigma}) 
\end{split}\nonumber\\
\begin{split}
\phantom{=}
\frac{1}{(\sqrt{2 \pi} \sigma)^N} \, e^{-(x_1^2 + x_2^2 + \dots + x_N^2)/2s^2\sigma^2} 
\frac{1}{\sigma^{\alpha_1 + \alpha_2 + \dots + \alpha_N} } 
\end{split}\nonumber\\
\begin{split}
=C \, (s \sigma)^{\alpha_1 + \alpha_2 + \dots + \alpha_N}
(-1)^{\alpha_1 + \alpha_2 + \dots + \alpha_N}
\end{split}\nonumber\\
\begin{split}
\phantom{=}
He_{\alpha_1}(\frac{x_1}{s\sigma}) \,  He_{\alpha_2}(\frac{x_2}{s\sigma}) \dots He_{\alpha_N}(\frac{x_N}{s\sigma}) 
\end{split}\nonumber\\
\begin{split}
\phantom{=}
\frac{1}{(\sqrt{2 \pi} s \sigma)^N} \, e^{-(x_1^2 + x_2^2 + \dots + x_N^2)/2s^2\sigma^2} 
\frac{1}{(s \sigma)^{\alpha_1 + \alpha_2 + \dots + \alpha_N} }.
\end{split}
\end{align}
Comparing with (\ref{eq-gauss-der-herm-pol}), we recognize this
expression as the scale-normalised derivative
\begin{equation}
h_s(x) = C \, (s \sigma)^{\alpha_1 + \alpha_2 + \dots + \alpha_N}
g_{x_1^{\alpha_1} x_2^{\alpha_2} \dots x_N^{\alpha_N}}(x;\; s \sigma)
\end{equation}
of order $\alpha = (\alpha_1, \alpha_2, \dots \alpha_N)$ at scale $s
\sigma$.

This means that if the scale channel network learns a partial Gaussian
derivative of some order, then the
application of that filter to all the scale channels is computationally equivalent to \emph{applying corresponding
scale-normalised Gaussian derivatives} to the original image at all
scales, given the complementary scaling transformation (\ref{eq:l_to_m})  with its
associated inverse mapping of
the scale parameters $s \mapsto s^{-1}$.

Specifically, this result implies that a scale channel network that
combines the multiple scale channels by a max pooling operation over
scales will have a similar function as scale selection performed by
detecting global extrema of scale-normalised derivatives over scales,
and thus share similarities to classical methods for scale selection
\cite{Lin97-IJCV,Lin14-EncCompVis}.

While this result has been expressed for partial derivatives, a
corresponding results holds also for derivative operators that
correspond to directional derivatives of Gaussian kernels in arbitrary
directions.
This result can be easily understood from the expression for a
directional derivative operator $\partial_{e^n}$ of order 
$n = n_1 + n_2 + \dots + n_N$ 
in direction $e = (e_1, e_2, \dots, e_N)$ with 
$|e| = \sqrt{e_1^2 + e_2^2 + \dots + e_N^2} = 1$:
\begin{align}
\begin{split}
  \partial_{e^n} g(x;\; \sigma) 
\end{split}\nonumber\\
\begin{split}
= (e_1 \, \partial_{x_1} + e_2 \, \partial_{x_2} + \dots +  e_{N} \, \partial_{x_N})^n g(x;\; \sigma)
\end{split}\nonumber\\
\begin{split}
= \sum_{\alpha_1 + \alpha_2 + \dots + \alpha_N = n}
{n \choose \alpha_1! \, \alpha_2! \, \dots \, \alpha_N!} 
\end{split}\nonumber\\
\begin{split}
\quad\quad\quad\quad\quad\quad
e_1^{\alpha_1} e_2^{\alpha_2} \dots e_N^{\alpha_N} \,
\partial_{x_1}^{\alpha_1} \partial_{x_2}^{\alpha_2} \dots \partial_{x_N}^{\alpha_N} 
g(x;\; \sigma)
\end{split}\nonumber\\
\begin{split}
= \sum_{\alpha_1 + \alpha_2 + \dots + \alpha_N = n}
{n \choose \alpha_1! \, \alpha_2! \, \dots \, \alpha_N!} 
\end{split}\nonumber\\
\begin{split}
\quad\quad\quad\quad\quad\quad
e_1^{\alpha_1} e_2^{\alpha_2} \dots e_N^{\alpha_N} \,
g_{x_1^{\alpha_1} x_2^{\alpha_2} \dots x_N^{\alpha_N}}(x;\; \sigma).
\end{split}
\end{align}
Since the scale normalisation factors $\sigma^{|\alpha|}$
%$\sigma^{\alpha_1 + \alpha_2 + \dots +\alpha_N}$ 
for all scale-normalised partial derivatives
of the same order $|\alpha| = \alpha_1 + \alpha_2 + \dots + \alpha_N = n$ are the same, it follows that all linear
combinations of partial derivatives of the same order are 
transformed by the same multiplicative scale normalisation factor, which
proves the result.

%\subsubsection{Scale channel networks and scale-normalised derivatives}
%Classical methods for scale selection detect maxima over scale of scale-normalised Gaussian derivatives \cite{lindeberg97-IJCV,scaleselection-Lin14}. We, here show that if a scale channel network which learns a single size filter corresponding to a Gaussian derivative, the equivalent kernels in $L_h$ correspond to such scale-normalised derivatives.
%
%Thus, (continuous) scale channel networks are a non-linear generalisation of the linear Gaussian scale space \emph{in the sense that} these networks can express the same primitive features which arise from the Gaussian scale space framework, but also multi-scale representations that are built upon more complex non-linear features. Since deep network tends to learn filters that look similar to Gaussian derivatives in the first layers one might expect this to be a good model for learning a more complex multi-scale representations from data. 
%
%A difference is that the features learned in the scale space network are clearly not guaranteed to correspond to gradual simplification of images (i.e. to fulfill the criteria of non-enhancement of local extrema). 

\begin{figure*}[hbpt]
	\begin{center}
		\includegraphics[width=1.0\textwidth]{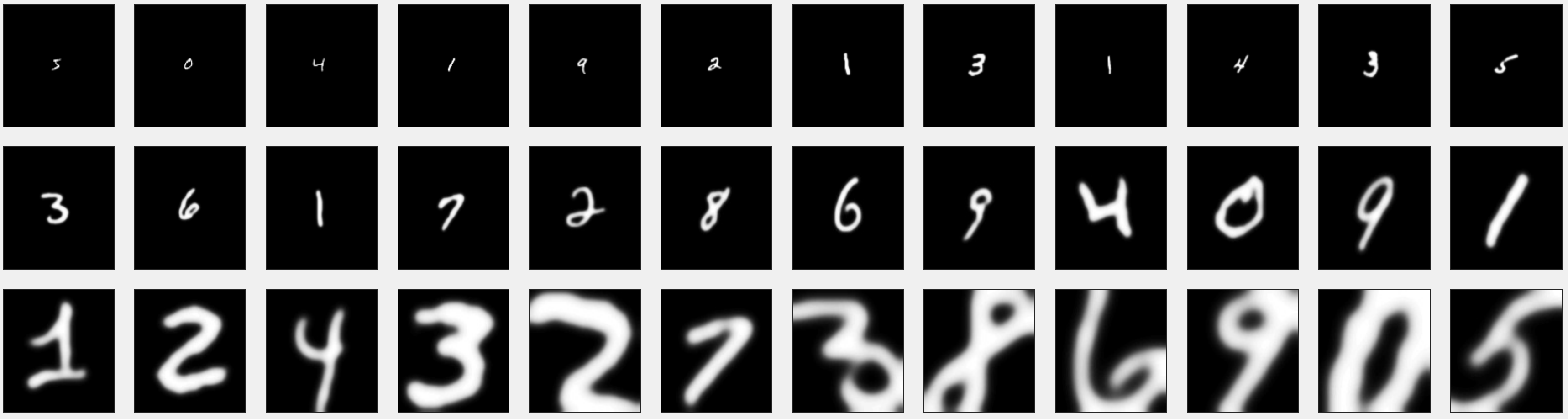} 
	\end{center}
	\caption{\emph{Samples from the MNIST Large Scale dataset}. The MNIST Large Scale dataset is derived from the original MNIST dataset \cite{MNISToriginal-1998} and contains $112 \times 112$ sized images of handwritten digits with scale variations of a factor of 16. The scale factors relative the original MNIST dataset are in the range $\frac{1}{2}$ (top left) to $8$ (bottom right).}
	\label{fig:dataset-mnist-scale}
\end{figure*}

\subsection{The MNIST Large Scale dataset}
	\label{app:mnist-large-scale}
	We, here, give a more detailed description of the \emph{MNIST Large Scale dataset}. The original MNIST dataset \cite{MNISToriginal-1998} contains $28\times28$ resolution images of centered handwritten digits. The MNIST Large Scale  dataset is derived from the MNIST dataset by rescaling the original MNIST images. The resulting dataset contains $112 \times 112$ resolution images with scale variations of a factor of $16$. The scale factors $s$ relative the original MNIST images are $s \in [\frac{1}{2}, 8]$. The dataset is illustrated in Figure \ref{fig:dataset-mnist-scale}.
	
% embedding the rescaled images in a $112x112$ resolution image. The resulting  dataset contains scale variations of a factor of $16$, where the scale factors  $s$ relative the original MNIST images is $s \in [0.5, 8]$. 
	To create an image with a certain scale factor $s$, the original image is first rescaled/resampled using bicubic interpolation. The image range is then clipped to $[0, 256]$ to remove possible over/undershoot resulting from the bicubic interpolation. The resulting image is embedded into an $112 \times 112$ resolution image using zero padding or cropping as needed. 
	% using the matlab affine2d + imwarp function?
	
	Large amounts of upsampling tends to result in discretisation artefacts. To reduce the severity of such artefacts, the images are post-processed with discrete Gaussian smoothing \cite{Lin90-PAMI} followed by non-linear thresholding. The standard deviation of the discrete Gaussian kernel varies with the scale factor as $\sigma(s) = \frac{7}{8}s$. After smoothing, the image range is rescaled to the range $[0, 255]$.
	%In the smoothing step, the standard deviation of the discrete Gaussian kernel  is $\sigma(s) = \frac{7}{8}s$.	
	%, where $s$ is the scaling factor.
	
 As a final step, an $\arctan$ non-linearity is applied to sharpen the resulting image, where
	the final image intensity $I_{out}$ is computed from the output of the smoothing step $I_{in}$ as:  %sharpening step 
	\begin{equation}
	I_{out} = \frac{2}{\pi}\arctan(a(I_{in} - b))
	\end{equation}	
	with $a=0.02$  and $b = 128$. 
%	The scaled images are created by resampling the original MNIST images using bicubic interpolation followed by smoothing and soft thresholding to reduce discretization effects. Smoothing is done with a discrete scale space kernel [ref?] where the standard deviation depends on the scale factor as $4/128$x times the size of the image afetr rescaling before padding correspoding to a smoothing factor of $\frac{7}{8}$ for the original image size (28x28 pixels).

	Note that for scale factors $>4$, the full digit might not be visible in the image. These scale factors are included to enable studying the limits of generalisation when the entire object is no longer visible (typically the digits are fully contained in the image for $s < 4\sqrt{2}$).
	% starts to be clear at scale factors larger than $s > 4\sqrt(2)$). 
	
	All training data sets are created from the first 50\,000 images in the original MNIST training set, while the last 10\,000 images in the original MNIST training set are used to create validation sets. The testing data sets are created by rescaling the 10\,000 images in the original MNIST test set. For the multi-scale datasets, scale factors for the individual images are sampled uniformly on a logarithmic scale in the range $[s_{min}, s_{max}]$.
	
	The specific datasets used for the experiments in this paper are available online \cite{MnistLargeScale-online}.	
	
\subsection{Numerical performance scores}
\label{app:numerical}
We here present the numerical performance scores for the experiments performed in this paper. The performance scores for (i) generalisation ability to unseen scales (Figure \ref{fig:single-scale-generalisation}) and (ii) learning in the presence of large scale variability (Figure \ref{fig:compare_multiscale_all}) are given in Table \ref{tab:benchmark-single-multi-scale}. The performance scores when (iii) varying the distance between consecutive scale channels for the FovMax and FovAvg networks (Figure \ref{fig:denseness}) are given in Table \ref{tab:denseness}. The performance scores for (iv) learning from small training sets with large scale variability (Figure \ref{fig:mnist_few_samples}) are given in Table \ref{tab:benchmark-few-samples}. When evaluating how the performance varies with the number of training samples, the first $n$ samples from the training set are used for training, while the full test set is used for testing.
%
%is given Table X.
%\begin{enumerate}[(i)]
%	\item Generalisation ability to unseen scales (Figure \ref{fig:single-scale-generalisation}) is given Table X.
%	\item Performance vs scale channel density for the FovMax and FovAvg network (Figure ) is given in Table X.
%	\item Learning in the presence of large scale varibility  in Table X.
%	\item (iv) Learning from small training sets with large scale variability (Figure \ref{fig:mnist_few_samples}) in Table X.
%\end{enumerate}
%
%
%We here present the numerical performance scores for the different experiments in the paper:
%\begin{enumerate}[(i)]
%\item Generalisation ability to unseen scales (Figure \ref{fig:single-scale-generalisation}) is given Table X.
%\item Performance vs scale channel density for the FovMax and FovAvg network (Figure ) is given in Table X.
%\item Learning in the presence of large scale varibility (Figure \ref{fig:compare_multiscale_all}) in Table X.
%\item (iv) Learning from small training sets with large scale variability (Figure \ref{fig:mnist_few_samples}) in Table X.
%\end{enumerate}
%
%The numerical scores for Figure \ref{fig:single-scale-generalisation} and Figure \ref{fig:compare_multiscale_all} are given in Table \ref{tab:benchmark-single-multi-scale}. The numerical scores for Figure \ref{fig:compare_multiscale_all} are given in Table \ref{tab:benchmark-few-samples}.

\setlength{\tabcolsep}{3.6pt} % Default value: 6pt

\begin{table*}[tbh]
	\caption{\emph{Classification accuracy (\%) as a function of the test set scale factor when training on single and multi-scale training data.} The table shows the performance for the different network architectures when trained on single scale training data (Figure \ref{fig:single-scale-generalisation}) or multi-scale training data (Figure \ref{fig:compare_multiscale_all}) from the MNIST Large Scale data set. 
	The networks are trained on either single scale data of scale 1, 2 or 4 (tr1, tr2, tr4) or multi-scale training data spanning the scale range $[1,4]$ (tr14). The FovMax, FovAvg, and SWMax networks all have 17 scale channels spanning the scale range $[\frac{1}{2}, 8]$. The FovConc networks have either 3 scale channels spanning the scale range $[1, 4]$ (3ch) or 5 scale channels spanning the scale range $[\frac{1}{2}, 8]$ (5ch), since using fewer scale channels improves the performance of this architecture in the setting with novel scale factors in the test set.
}
	\begin{tabular}{l r r r r r r r r r r r r r r r r r }

%Scales & 0.50 & 0.59 & 0.71 & 0.84 & 1.00 & 1.19 & 1.41 & 1.68 & 2.00 & 2.38 & 2.83 & 3.36 & 4.00 & 4.76 & 5.66 & 6.73 & 8.00 \\ 
\hline
Scales & $1/2$ & $2^{-3/4}$ & $2^{-1/2}$ & $2^{-1/4}$ & 1 & $2^{1/4}$ & $2^{1/2}$ & $2^{3/4}$ & 2 & $2^{5/4}$ & $2^{3/2}$ & $2^{7/4}$ & 4 & $2^{9/4}$ & $2^{5/2}$ & $2^{11/4}$ & 8 \\ 
\hline
CNN tr1 & 61.84 & 85.31 & 96.10 & 98.73 & 99.32 & 98.50 & 85.36 & 52.61 & 36.82 & 28.55 & 22.38 & 19.04 & 14.47 & 11.71 & 11.50 & 10.88 & 10.68 \\ 
CNN tr2 & 14.37 & 15.81 & 22.17 & 33.42 & 53.57 & 80.70 & 95.52 & 98.87 & 99.38 & 98.81 & 90.31 & 60.95 & 40.85 & 29.91 & 23.69 & 19.26 & 16.68 \\ 
CNN tr4 & 9.89 & 10.41 & 11.33 & 12.06 & 13.57 & 15.96 & 18.54 & 25.97 & 44.93 & 76.62 & 95.19 & 98.96 & 99.40 & 98.71 & 92.99 & 74.21 & 47.63 \\ 
CNN tr14 & 67.34 & 84.02 & 93.71 & 97.38 & 98.70 & 99.12 & 99.23 & 99.29 & 99.34 & 99.32 & 99.31 & 99.05 & 98.45 & 95.40 & 81.25 & 56.65 & 38.17 \\ 
\hline
FovAvg 17ch tr1 & 98.58 & 99.05 & 99.33 & 99.39 & 99.40 & 99.39 & 99.38 & 99.36 & 99.35 & 99.31 & 99.22 & 99.12 & 98.94 & 98.47 & 96.20 & 89.17 & 71.31 \\ 
FovAvg 17ch tr2 & 98.58 & 99.04 & 99.36 & 99.35 & 99.38 & 99.37 & 99.37 & 99.37 & 99.38 & 99.35 & 99.36 & 99.34 & 99.32 & 99.25 & 98.83 & 96.89 & 88.46 \\ 
FovAvg 17ch tr4 & 97.65 & 98.68 & 99.14 & 99.17 & 99.28 & 99.26 & 99.28 & 99.27 & 99.30 & 99.32 & 99.35 & 99.37 & 99.39 & 99.34 & 98.90 & 96.75 & 88.68 \\ 
FovAvg 17ch tr14 & 98.78 & 99.17 & 99.30 & 99.37 & 99.40 & 99.40 & 99.40 & 99.40 & 99.41 & 99.40 & 99.39 & 99.40 & 99.39 & 99.36 & 99.05 & 96.55 & 88.17 \\ 
\hline
FovMax 17ch tr1 & 98.71 & 99.07 & 99.27 & 99.34 & 99.37 & 99.35 & 99.36 & 99.34 & 99.33 & 99.35 & 99.34 & 99.35 & 99.34 & 99.27 & 97.88 & 92.76 & 79.23 \\ 
FovMax 17ch tr2 & 98.75 & 99.12 & 99.25 & 99.30 & 99.34 & 99.29 & 99.31 & 99.32 & 99.32 & 99.32 & 99.30 & 99.32 & 99.32 & 99.12 & 97.43 & 91.87 & 75.85 \\ 
FovMax 17ch tr4 & 98.49 & 98.97 & 99.18 & 99.25 & 99.28 & 99.31 & 99.29 & 99.30 & 99.29 & 99.30 & 99.30 & 99.30 & 99.31 & 99.28 & 98.33 & 92.39 & 77.01 \\ 
FovMax 17ch tr14 & 98.71 & 99.13 & 99.30 & 99.31 & 99.35 & 99.31 & 99.32 & 99.31 & 99.32 & 99.33 & 99.32 & 99.32 & 99.32 & 99.24 & 98.40 & 93.46 & 81.42 \\ 
\hline
FovConc 3ch tr1 & 40.38 & 71.00 & 94.34 & 98.83 & 99.28 & 98.61 & 89.29 & 60.43 & 36.36 & 22.91 & 15.29 & 9.74 & 5.86 & 3.83 & 3.21 & 4.14 & 6.02 \\ 
FovConc 3ch tr2 & 9.47 & 10.11 & 16.84 & 28.67 & 46.66 & 74.87 & 94.76 & 98.88 & 99.35 & 98.55 & 89.35 & 61.24 & 39.84 & 27.52 & 18.44 & 14.68 & 13.08 \\ 
FovConc 3ch tr4 & 5.49 & 3.68 & 1.29 & 1.08 & 1.29 & 2.73 & 9.23 & 23.47 & 45.24 & 76.25 & 95.25 & 98.98 & 99.35 & 98.55 & 91.51 & 72.29 & 50.07 \\ 
FovConc 5ch tr2 & 1.84 & 2.66 & 7.89 & 20.96 & 35.58 & 65.31 & 93.16 & 98.71 & 99.23 & 98.45 & 88.55 & 47.87 & 17.48 & 6.96 & 4.13 & 4.09 & 6.21 \\ 
FovConc 5ch tr14 & 70.45 & 91.66 & 98.00 & 98.95 & 99.22 & 99.26 & 99.26 & 99.25 & 99.25 & 99.25 & 99.23 & 99.22 & 99.19 & 98.91 & 96.89 & 88.77 & 72.05 \\ 
\hline
SWMax 17ch tr1 & 84.58 & 95.23 & 98.10 & 98.59 & 98.78 & 98.51 & 98.49 & 98.39 & 98.35 & 98.10 & 97.68 & 96.67 & 93.48 & 84.46 & 70.31 & 56.19 & 43.17 \\ 
SWMax 17ch tr2 & 91.86 & 96.82 & 98.28 & 98.63 & 98.77 & 98.75 & 98.67 & 98.74 & 98.88 & 98.64 & 98.49 & 97.55 & 93.14 & 85.67 & 71.52 & 54.61 & 41.47 \\ 
SWMax 17ch tr4 & 78.60 & 91.58 & 94.54 & 95.46 & 96.83 & 96.73 & 95.88 & 95.36 & 97.07 & 97.74 & 98.12 & 98.64 & 98.74 & 97.84 & 89.27 & 71.95 & 53.28 \\ 
SWMax 17ch tr14 & 80.08 & 89.98 & 93.81 & 95.36 & 96.53 & 96.67 & 95.83 & 95.62 & 96.53 & 96.99 & 97.47 & 98.15 & 98.55 & 97.14 & 85.27 & 64.82 & 44.18 \\ 
\hline
	\end{tabular} 
\label{tab:benchmark-single-multi-scale}
\end{table*}

\setlength{\tabcolsep}{3.2pt} % Default value: 6pt

\begin{table*}[tbh]
	\caption{\emph{Classification accuracy (\%) as a function of the test set scale factor for the FovMax and FovAvg networks when varying the distance between consecutive scale channels}. 
		The table shows the performance for different test set scale factors when training the FovMax and FovAvg networks spanning the scale range $[\frac{1}{4},8]$ 
		with varying distance between consecutive scale channels ($2^{1/4}$, $2^{1/2}$, $2$). 
		All networks are trained on single scale training data of scale 2 from the MNIST Large Scale data set. 
		%	Varying the sampling density of the scale channels. FovMax and FovAvg networks spanning the scale range $[\frac{1}{4},8]$ are trained with varying spacing between the scale channels ($2$, $2^{1/2}$ and $2^{1/4}$). All networks are trained on scale 2 training data. There is a significant increase in the performance when reducing the spacing between the scale channels from $2$ to $2^{1/2}$ while the effect of a further reduction to $2^{1/4}$ is small. 
	}
	\begin{tabular}{l r r r r r r r r r r r r r r r r r } 
%Scales & 0.50 & 0.59 & 0.71 & 0.84 & 1.00 & 1.19 & 1.41 & 1.68 & 2.00 & 2.38 & 2.83 & 3.36 & 4.00 & 4.76 & 5.66 & 6.73 & 8.00 \\ 
		\hline
	Scales & $1/2$ & $2^{-3/4}$ & $2^{-1/2}$ & $2^{-1/4}$ & 1 & $2^{1/4}$ & $2^{1/2}$ & $2^{3/4}$ & 2 & $2^{5/4}$ & $2^{3/2}$ & $2^{7/4}$ & 4 & $2^{9/4}$ & $2^{5/2}$ & $2^{11/4}$ & 8  \\ 
		\hline
		FovMax $2^{1/4}$ (17ch)& 98.75 & 99.12 & 99.25 & 99.30 & 99.34 & 99.29 & 99.31 & 99.32 & 99.32 & 99.32 & 99.30 & 99.32 & 99.32 & 99.12 & 97.43 & 91.87 & 75.85 \\ 
		FovMax $2^{1/2}$ (9ch)& 98.51 & 99.03 & 99.26 & 99.34 & 99.31 & 99.36 & 99.32 & 99.37 & 99.32 & 99.35 & 99.29 & 99.35 & 99.30 & 99.22 & 97.40 & 91.59 & 77.75 \\ 
		FovMax $2$ (5ch)& 97.45 & 96.44 & 97.32 & 98.77 & 99.21 & 98.57 & 97.58 & 98.74 & 99.20 & 98.64 & 97.51 & 98.71 & 99.20 & 98.59 & 93.59 & 83.55 & 69.20 \\ 
		\hline
		FovAvg $2^{1/4}$ (17ch)& 98.66 & 99.11 & 99.35 & 99.38 & 99.40 & 99.41 & 99.43 & 99.42 & 99.42 & 99.41 & 99.42 & 99.40 & 99.38 & 99.31 & 98.95 & 96.76 & 88.38 \\ 
		FovAvg $2^{1/2}$ (9ch)& 98.75 & 99.07 & 99.26 & 99.34 & 99.36 & 99.36 & 99.36 & 99.39 & 99.35 & 99.37 & 99.35 & 99.38 & 99.31 & 99.31 & 98.73 & 96.67 & 88.44 \\ 
		FovAvg $2$ (5ch) & 98.84 & 98.74 & 98.81 & 99.18 & 99.36 & 99.15 & 98.85 & 99.23 & 99.33 & 99.12 & 98.84 & 99.20 & 99.31 & 99.06 & 97.70 & 94.20 & 86.20 \\ 
		\hline
	\end{tabular} 
	\label{tab:denseness}
\end{table*}

\setlength{\tabcolsep}{4.5pt} % Default value: 6pt

\begin{table}[tbh]
\caption{\emph{Classification accuracy (\%) as a function of the number of training samples when training on multi-scale training data}. The table shows the performance of the different network architectures when trained with a gradually reduced number of training samples from the MNIST Large Scale dataset (Figure \ref{fig:mnist_few_samples}). Both the training and test sets here span the scale range $[1,4]$. The FovAvg, FovMax, FovConc and SWMax networks all have 17 scale channels spanning the scale range $[\frac{1}{2}, 8]$.}
%The FovAvg network shows the highest robustness when decreasing the number of training samples followed by the FovMax network. 
\begin{center}
\begin{tabular}{l r r r r r r r}
\hline
\# samples & 10\,000 & 5\,000 & 2\,000 & 1\,000 & 500 & 200 & 100 \\ 
\hline
CNN & 97.95 & 96.95 & 94.71 & 92.01 & 87.85 & 71.85 & 52.48 \\ 
FovAvg 17ch & 98.76 & 98.35 & 96.93 & 96.61 & 94.73 & 87.56 & 80.89 \\ 
FovMax 17ch & 98.56 & 97.89 & 96.66 & 93.98 & 90.10 & 79.26 & 68.24 \\
%FovConc 17ch & 98.56 & 97.62 & 96.45 & 94.36 & 89.12 & 7	5.89 & 59.68 \\ 
FovConc 17ch & 98.49 & 97.96 & 96.43 & 94.52 & 89.37 & 75.48 & 60.40 \\
SWMax & 97.55 & 96.11 & 92.90 & 86.40 & 80.91 & 62.83 & 49.36 \\ 
\hline
\end{tabular}
\end{center}
\label{tab:benchmark-few-samples}
\end{table}

\bibliographystyle{IEEEtran}
\bibliography{defs,tlmac,yjrefs,yjdeepl,yjclassic}

\begin{thebibliography}{10}
\providecommand{\url}[1]{#1}
\csname url@samestyle\endcsname
\providecommand{\newblock}{\relax}
\providecommand{\bibinfo}[2]{#2}
\providecommand{\BIBentrySTDinterwordspacing}{\spaceskip=0pt\relax}
\providecommand{\BIBentryALTinterwordstretchfactor}{4}
\providecommand{\BIBentryALTinterwordspacing}{\spaceskip=\fontdimen2\font plus
\BIBentryALTinterwordstretchfactor\fontdimen3\font minus
  \fontdimen4\font\relax}
\providecommand{\BIBforeignlanguage}[2]{{%
\expandafter\ifx\csname l@#1\endcsname\relax
\typeout{** WARNING: IEEEtran.bst: No hyphenation pattern has been}%
\typeout{** loaded for the language `#1'. Using the pattern for}%
\typeout{** the default language instead.}%
\else
\language=\csname l@#1\endcsname
\fi
#2}}
\providecommand{\BIBdecl}{\relax}
\BIBdecl

\bibitem{BruMal-TPAMI2013}
J.~Bruna and S.~Mallat, ``Invariant scattering convolution networks,''
  \emph{IEEE Transactions on Pattern Analysis and Machine Intelligence},
  vol.~35, no.~8, pp. 1872--1886, 2013.

\bibitem{CohWel-ICML2016}
T.~Cohen and M.~Welling, ``Group equivariant convolutional networks,'' in
  \emph{International Conference on Machine Learning}, 2016, pp. 2990--2999.

\bibitem{LapSavBuh-CVPR2016}
D.~Laptev, N.~Savinov, J.~M. Buhmann, and M.~Pollefeys, ``{TI}-pooling:
  transformation-invariant pooling for feature learning in convolutional neural
  networks,'' in \emph{Proc. Conference on Computer Vision and Pattern
  Recognition (CVPR)}, 2016, pp. 289--297.

\bibitem{Iij62-TR}
T.~Iijima, ``Observation theory of two-dimensional visual patterns,'' Papers of
  Technical Group on Automata and Automatic Control, IECE, Japan, Tech. Rep.,
  1962, (in Japanese).

\bibitem{Wit83}
A.~P. Witkin, ``Scale-space filtering,'' in \emph{Proc. 8th Int. Joint Conf.
  Art. Intell.}, Karlsruhe, Germany, Aug. 1983, pp. 1019--1022.

\bibitem{Koe84-BC}
J.~J. Koenderink, ``The structure of images,'' \emph{Biological Cybernetics},
  vol.~50, pp. 363--370, 1984.

\bibitem{lindeberg1993-scspbook}
T.~Lindeberg, \emph{Scale-Space Theory in Computer Vision}.\hskip 1em plus
  0.5em minus 0.4em\relax Springer, 1993.

\bibitem{lindeberg1994-scsparticle}
------, ``Scale-space theory: A basic tool for analyzing structures at
  different scales,'' \emph{Journal of Applied Statistics}, vol.~21, no. 1-2,
  pp. 225--270, 1994.

\bibitem{Flo97-book}
L.~M.~J. Florack, \emph{Image Structure}, ser. Series in Mathematical Imaging
  and Vision.\hskip 1em plus 0.5em minus 0.4em\relax Springer, 1997.

\bibitem{scalespaceJapan_weickert99}
J.~Weickert, S.~Ishikawa, and A.~Imiya, ``Linear scale-space has first been
  proposed in {J}apan,'' \emph{Journal of Mathematical Imaging and Vision},
  vol.~10, no.~3, pp. 237--252, 1999.

\bibitem{haarromeny-04book}
B.~ter Haar~Romeny, \emph{Front-End Vision and Multi-Scale Image
  Analysis}.\hskip 1em plus 0.5em minus 0.4em\relax Berlin/New York: Springer,
  2003.

\bibitem{EngTsiSchMad-ICML2019}
L.~Engstrom, B.~Tran, D.~Tsipras, L.~Schmidt, and A.~Madry, ``Exploring the
  landscape of spatial robustness,'' in \emph{International Conference on
  Machine Learning (ICML)}, 2019, pp. 1802--1811.

\bibitem{FawFro-BMVC2015}
A.~Fawzi and P.~Frossard, ``Manitest: Are classifiers really invariant?''
  \emph{British Machine Vision Conference (BMVC)}, 2015.

\bibitem{SinDav-CVPR2018}
B.~Singh and L.~S. Davis, ``An analysis of scale invariance in object detection
  {SNIP},'' in \emph{Proc. Conference on Computer Vision and Pattern
  Recognition (CVPR)}, 2018, pp. 3578--3587.

\bibitem{CaiFanFerVas-ECCV2016}
Z.~Cai, Q.~Fan, R.~S. Feris, and N.~Vasconcelos, ``A unified multi-scale deep
  convolutional neural network for fast object detection,'' in \emph{European
  Conference on Computer Vision}.\hskip 1em plus 0.5em minus 0.4em\relax
  Springer, 2016, pp. 354--370.

\bibitem{JadSimZisKav-NIPS2015}
M.~Jaderberg, K.~Simonyan, A.~Zisserman, and K.~Kavukcuoglu, ``Spatial
  transformer networks,'' in \emph{Advances in Neural Information Processing
  Systems (NIPS)}, 2015, pp. 2017--2025.

\bibitem{YuKol-ICLR2016}
F.~Yu and V.~Koltun, ``Multi-scale context aggregation by dilated
  convolutions,'' in \emph{Int. Conf. on Learning Representations (ICLR)},
  2016.

\bibitem{SerEigZha-arXiv2013}
P.~Sermanet, D.~Eigen, X.~Zhang, M.~Mathieu, R.~Fergus, and Y.~LeCun,
  ``{OverFeat}: Integrated recognition, localization and detection using
  convolutional networks,'' \emph{arXiv preprint arXiv:1312.6229}, 2013.

\bibitem{Gir-ICCV2015}
R.~Girshick, ``Fast {R-CNN},'' in \emph{Proc. International Conference on
  Computer Vision (ICCV)}, 2015, pp. 1440--1448.

\bibitem{LinDolGirhe-CVPR2017}
T.-Y. Lin, P.~Doll{\'a}r, R.~Girshick, K.~He, B.~Hariharan, and S.~Belongie,
  ``Feature pyramid networks for object detection,'' in \emph{Proc. Conference
  on Computer Vision and Pattern Recognition}, 2017, pp. 2117--2125.

\bibitem{SifMal-CVPR2013}
L.~Sifre and S.~Mallat, ``Rotation, scaling and deformation invariant
  scattering for texture discrimination,'' in \emph{Proc. Conference on
  Computer Vision and Pattern Recognition (CVPR)}, 2013, pp. 1233--1240.

\bibitem{Lin2020provably}
T.~Lindeberg, ``Provably scale-covariant continuous hierarchical networks based
  on scale-normalized differential expressions coupled in cascade,''
  \emph{Journal of Mathematical Imaging and Vision}, vol.~62, no.~1, pp.
  120--148, 2020.

\bibitem{XuXiaZhaYan-arXiv2014}
Y.~Xu, T.~Xiao, J.~Zhang, K.~Yang, and Z.~Zhang, ``Scale-invariant
  convolutional neural networks,'' \emph{arXiv preprint arXiv:1411.6369}, 2014.

\bibitem{KanShaJac-arXiv2014}
A.~Kanazawa, A.~Sharma, and D.~W. Jacobs, ``Locally scale-invariant
  convolutional neural networks,'' \emph{arXiv preprint arXiv:1412.5104}, 2014.

\bibitem{MarKelLob-arXiv2018}
D.~Marcos, B.~Kellenberger, S.~Lobry, and D.~Tuia, ``Scale equivariance in
  {CNNs} with vector fields,'' \emph{arXiv preprint arXiv:1807.11783}, 2018.

\bibitem{WorWel-NIPS2019}
D.~Worrall and M.~Welling, ``Deep scale-spaces: Equivariance over scale,'' in
  \emph{Advances in Neural Information Processing Systems (NIPS)}, 2019, pp.
  7364--7376.

\bibitem{FarCouNajLeC-2013}
C.~Farabet, C.~Couprie, L.~Najman, and Y.~LeCun, ``Learning hierarchical
  features for scene labeling,'' \emph{IEEE Transactions on Pattern Analysis
  and Machine Intelligence}, vol.~35, no.~8, pp. 1915--1929, 2013.

\bibitem{NooPos-PR2016}
N.~Van~Noord and E.~Postma, ``Learning scale-variant and scale-invariant
  features for deep image classification,'' \emph{Pattern Recognition},
  vol.~61, pp. 583--592, 2017.

\bibitem{CVAP166}
T.~Lindeberg and L.~Florack, ``Foveal scale-space and linear increase of
  receptive field size as a function of eccentricity,'' Dept. of Numerical
  Analysis and Computer Science, KTH, report ISRN KTH/NA/P-{}-94/27-{}-SE, Aug.
  1994, available from https://people.kth.se/\texttildelow
  tony/papers/cvap166.pdf.

\bibitem{lindeberg97-IJCV}
T.~Lindeberg, ``Feature detection with automatic scale selection,''
  \emph{International Journal of Computer Vision}, vol.~30, no.~2, pp. 77--116,
  1998.

\bibitem{scaleselection-Lin14}
------, ``Scale selection,'' in \emph{Computer Vision: A Reference Guide},
  K.~Ikeuchi, Ed.\hskip 1em plus 0.5em minus 0.4em\relax Springer, 2014, pp.
  701--713.

\bibitem{LiTaxLoo11-ScSp}
Y.~Li, D.~M.~J. Tax, and M.~Loog, ``Supervised scale-invariant segmentation
  (and detection),'' in \emph{Proc.\ Scale Space and Variational Methods in
  Computer Vision (SSVM 2011)}, ser. Springer LNCS, vol. 6667.\hskip 1em plus
  0.5em minus 0.4em\relax Ein Gedi, Israel: Springer, 2012, pp. 350--361.

\bibitem{LooLiTax09-LNCS}
M.~Loog, Y.~Li, and D.~Tax, ``Maximum membership scale selection,'' in
  \emph{Multiple Classifier Systems}, ser. Springer LNCS, vol. 5519, 2009, pp.
  468--477.

\bibitem{Lin15-JMIV}
T.~Lindeberg, ``Image matching using generalized scale-space interest points,''
  \emph{Journal of Mathematical Imaging and Vision}, vol.~52, no.~1, pp. 3--36,
  2015.

\bibitem{scaleselection-Lin13JMIV}
------, ``Scale selection properties of generalized scale-space interest point
  detectors,'' \emph{Journal of Mathematical Imaging and vision}, vol.~46,
  no.~2, pp. 177--210, 2013.

\bibitem{MNISToriginal-1998}
Y.~LeCun, L.~Bottou, Y.~Bengio, P.~Haffner \emph{et~al.}, ``Gradient-based
  learning applied to document recognition,'' \emph{Proc. of the IEEE},
  vol.~86, no.~11, pp. 2278--2324, 1998.

\bibitem{MnistLargeScale-online}
Y.~Jansson and T.~Lindeberg, ``{MNIST Large Scale} dataset,'' [Online].
  Available at: https://www.zenodo.org/record/3820247.
  {DOI}:10.5281/zenodo.3820247, 2020.

\bibitem{Lin97-IJCV}
T.~Lindeberg, ``Feature detection with automatic scale selection,''
  \emph{International Journal of Computer Vision}, vol.~30, no.~2, pp. 77--116,
  1998.

\bibitem{Lin98-IJCV}
------, ``Edge detection and ridge detection with automatic scale selection,''
  \emph{International Journal of Computer Vision}, vol.~30, no.~2, pp.
  117--154, 1998.

\bibitem{BL97-CVIU}
L.~Bretzner and T.~Lindeberg, ``Feature tracking with automatic selection of
  spatial scales,'' \emph{Computer Vision and Image Understanding}, vol.~71,
  no.~3, pp. 385--392, Sep. 1998.

\bibitem{ChoVerHalCro00-ECCV}
O.~Chomat, V.~de~Verdiere, D.~Hall, and J.~Crowley, ``Local scale selection for
  {G}aussian based description techniques,'' in \emph{Proc.\ European Conf. on
  Computer Vision (ECCV 2000)}, ser. Springer LNCS, vol. 1842, Dublin, Ireland,
  2000, pp. I:117--133.

\bibitem{MikSch04-IJCV}
K.~Mikolajczyk and C.~Schmid, ``Scale and affine invariant interest point
  detectors,'' \emph{International Journal of Computer Vision}, vol.~60, no.~1,
  pp. 63--86, 2004.

\bibitem{Low04-IJCV}
D.~G. Lowe, ``Distinctive image features from scale-invariant keypoints,''
  \emph{International Journal of Computer Vision}, vol.~60, no.~2, pp. 91--110,
  2004.

\bibitem{BayEssTuyGoo08-CVIU}
H.~Bay, A.~Ess, T.~Tuytelaars, and L.~van Gool, ``Speeded up robust features
  {(SURF)},'' \emph{Computer Vision and Image Understanding}, vol. 110, no.~3,
  pp. 346--359, 2008.

\bibitem{TuyMik08-Book}
T.~Tuytelaars and K.~Mikolajczyk, \emph{A Survey on Local Invariant Features},
  ser. Foundations and Trends in Computer Graphics and Vision.\hskip 1em plus
  0.5em minus 0.4em\relax Now Publishers, 2008, vol. 3(3).

\bibitem{Lin13-ImPhys}
T.~Lindeberg, ``Generalized axiomatic scale-space theory,'' in \emph{Advances
  in Imaging and Electron Physics}, P.~Hawkes, Ed.\hskip 1em plus 0.5em minus
  0.4em\relax Elsevier, 2013, vol. 178, pp. 1--96.

\bibitem{Lin14-EncCompVis}
------, ``Scale selection,'' in \emph{Computer Vision: A Reference Guide},
  K.~Ikeuchi, Ed.\hskip 1em plus 0.5em minus 0.4em\relax Springer, 2014, pp.
  701--713.

\bibitem{Lin90-PAMI}
------, ``Scale-space for discrete signals,'' \emph{IEEE Trans. Pattern
  Analysis and Machine Intell.}, vol.~12, no.~3, pp. 234--254, Mar. 1990.

\end{thebibliography}

\end{document}